\documentclass{article}
\usepackage[accepted]{icml2022}
\usepackage{fancyhdr}
\usepackage{framed,multirow}
\usepackage{natbib}
\usepackage{graphicx}
\usepackage{amssymb}
\usepackage{latexsym}

\usepackage{url}
\usepackage{xcolor}

\usepackage{algorithmic}
\usepackage{booktabs} 
\usepackage[colorlinks=false,linkcolor=blue]{hyperref}
\usepackage{bbm}
\usepackage{amsmath,amsthm,amsfonts,amssymb,amscd}
\usepackage{csvsimple}
\usepackage{xcolor}
\usepackage[normalem]{ulem} %
\usepackage{array}

\usepackage{csquotes}

\definecolor{blue}{RGB}{17, 85, 204}

\usepackage{makecell} 
\newcommand{\PreserveBackslash}[1]{\let\temp=\\#1\let\\=\temp}
\newcolumntype{C}[1]{>{\PreserveBackslash\centering}p{#1}}
\newcolumntype{R}[1]{>{\PreserveBackslash\raggedleft}p{#1}}
\newcolumntype{L}[1]{>{\PreserveBackslash\raggedright}p{#1}}
\definecolor{newcolor}{rgb}{.8,.349,.1}
\usepackage{placeins}

\title{\textbf{Guidelines and Evaluation of Clinical Explainable AI in \\Medical Image Analysis}}

\author{Weina Jin$^1$ \and Xiaoxiao Li$^2$ \and Mostafa Fatehi$^3$ \and Ghassan Hamarneh$^1$}
\date{%
    $^1$School of Computing Science, Simon Fraser University\\%
    $^2$Department of Electrical and Computer Engineering, The University of British Columbia\\
    $^3$Division of Neurosurgery, The University of British Columbia\\[2ex]%
}

\begin{document}
\maketitle
\footnotetext[1]{The preprint version of the Medical Image Analysis article: \href{https://doi.org/10.1016/j.media.2022.102684}{https://doi.org/10.1016/j.media.2022.102684} . \\\copyright 2022. This manuscript version is made available under the CC-BY-NC-ND 4.0 license \href{https://creativecommons.org/licenses/by-nc-nd/4.0/}{https://creativecommons.org/licenses/by-nc-nd/4.0/}}

\begin{abstract}

Explainable artificial intelligence (XAI) is essential for enabling clinical users to get informed decision support from AI and comply with evidence-based medical practice. Applying XAI in clinical settings requires proper evaluation criteria to ensure the explanation technique is both technically sound and clinically useful, but specific support is lacking to achieve this goal.
To bridge the research gap, we propose the Clinical XAI Guidelines that consist of five criteria a clinical XAI needs to be optimized for. The guidelines recommend choosing an explanation form based on Guideline 1 (G1) Understandability and G2 Clinical relevance. For the chosen explanation form, its specific XAI technique should be optimized for G3 Truthfulness, G4 Informative plausibility, and G5 Computational efficiency.
Following the guidelines, we conducted a systematic evaluation on a novel problem of multi-modal medical image explanation with two clinical tasks, and proposed new evaluation metrics accordingly. Sixteen commonly-used heatmap XAI techniques were evaluated and found to be insufficient for clinical use due to their failure in G3 and G4. Our evaluation demonstrated the use of Clinical XAI Guidelines to support the design and evaluation of clinically viable XAI.

\end{abstract}

\section{Introduction}\label{intro}
\begin{figure*}[!ht]
    \centering    \includegraphics[width=1\linewidth]{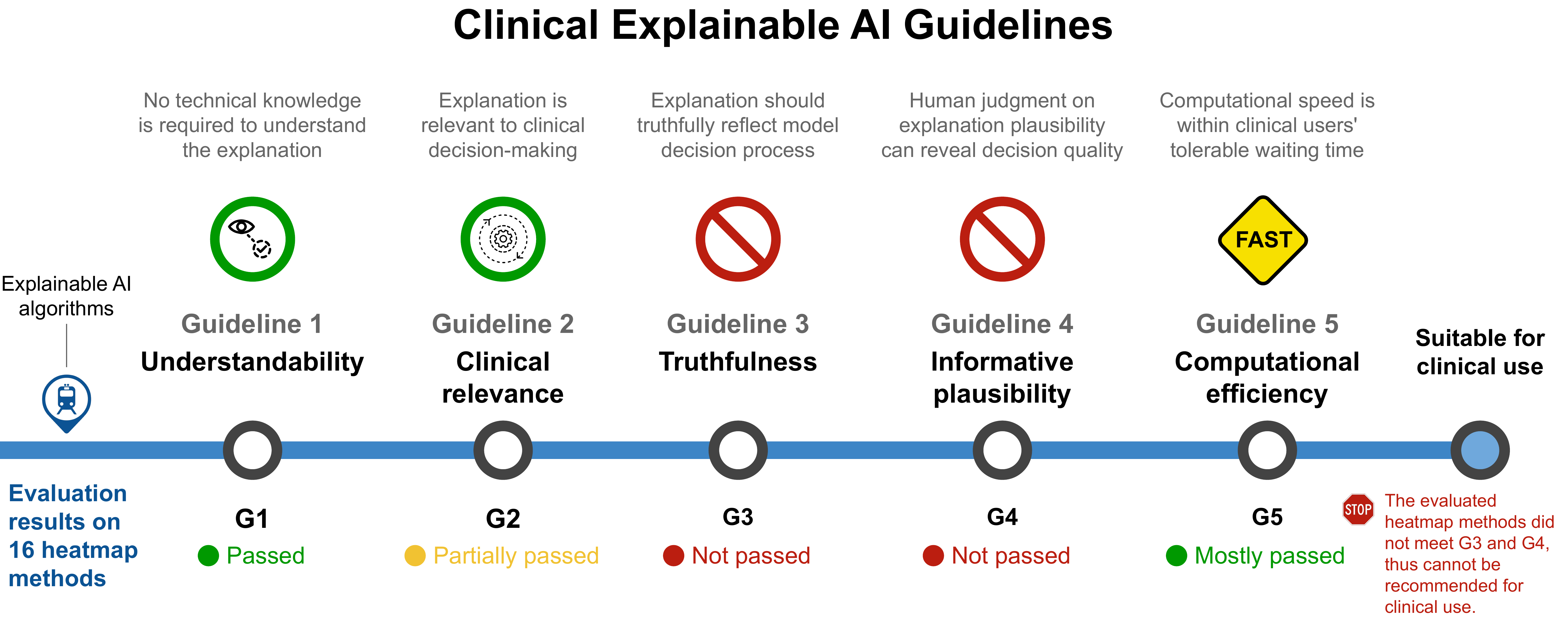}
    \caption{The Clinical Explainable AI Guidelines. Explainable AI algorithms should meet the five criteria in the guideline to be suitable for clinical use. The evaluation results on 16 heatmap methods regarding the guidelines criteria are shown at the bottom.}
    \label{fig:gl}
\end{figure*}
Suppose an artificial intelligence (AI) developer Alex is developing a clinical AI system, and she wants to select an explainable AI (XAI) technique to make the AI model interpretable and transparent to clinical users. As there are numerous AI explainability techniques available, Alex may ask: \textit{How can I choose an AI explainability technique that is optimal for my target clinical task?} She may look up literature on XAI evaluation~\citep{Sokol2020,10.1145/3387166,VILONE202189,8400040,DBLP:journals/corr/abs-1806-00069} hoping it will guide her selection on XAI techniques. The literature suggests various selection criteria and computational- or human-level evaluation methods. But since Alex is building an AI system which will assist doctors in clinically important decisions, she may ask,
\textit{Is it clinically viable to use these evaluation metrics? Will they help to meet doctors' clinical requirements \textcolor{black}{}for AI explanation? How to prioritize multiple evaluation objectives for clinical XAI systems? }

Alex's questions are prevalent when applying or proposing explainable AI techniques for clinical use. 
As a fast-advancing technology, AI has transformative potential in many medical fields%
~\citep{Zhang2019,Fujisawa2018,Mohan2020}. Nonetheless, there are outstanding barriers to the widespread translation of AI from bench to bedside~\citep{He2019}, such as data collection and harmonization~\citep{NAN202299}, data privacy~\citep{10.3389/frai.2021.746497}, bias and fairness in data and model~\citep{DBLP:journals/corr/abs-2110-00603,Rajpurkar2022}, domain adaptation and generalization~\citep{Futoma2020}, and model explainability%
~\citep{Jin_2020,Rajpurkar2022, Kelly2019}. In this work, we focus on the problem of AI model explainability, interpretability, or transparency. The model explainability issue is caused by the black-box nature of the state-of-the-art AI technologies, i.e., deep neural networks (DNN): the decision process of AI models is not completely and intuitively comprehensible even to its human creators, due to its millions of parameters, complex feature representations in high-dimensional space, multiple layers of decision processing, and non-linear mappings from input space to output prediction.

AI developers, like Alex, resort to XAI techniques to explain AI decisions in human-understandable forms~\citep{doshivelez2017rigorous}, and enable clinical users to make informed decisions
with AI assistance 
that comply with evidence-based medical practice\footnote{``Evidence-based medicine is the conscientious, explicit, judicious, and reasonable use of modern, best evidence in making decisions about the care of individual patients.''~\citep{Masic2008}}~\citep{Sackett71}.
The notion of XAI and its corresponding techniques were originally proposed in the machine learning community~\citep{BARREDOARRIETA202082, 10.1145/3236009, Zhang2018a}, and were then applied and developed in the medical image analysis (MIA) community~\citep{YANG202229, Singh2020}, for example in brain ~\citep{PEREIRA2018228}, retinal~\citep{DeFauw2018}, cardiac~\citep{Bello2019}, chest~\citep{YE2022108291}, and skin imaging tasks~\citep{8333693}. They \textcolor{black}{}utilize different explanation forms and algorithms that aim to generate clinical end-user-friendly explanations~\citep{jin2021euca}, such as explaining using features (heatmap~\citep{Bien2018}, concept~\citep{pmlr-v80-kim18d}), examples (similar~\citep{10.1145/3290605.3300234}, typical~\citep{10.5555/3454287.3455088}, and counterfactual examples~\citep{10.1007/978-3-030-32226-7_76}), and rules (decision tree~\citep{Wu2019a}).
Indeed, research has shown that explanations have the potential to help clinical users to verify AI's decisions~\citep{Ribeiro2016b}, resolve disagreements with AI during decision discrepancy~\citep{10.1145/3359206}, calibrate their trust in AI assistance~\citep{7349687,Zhang2020}, identify potential biases~\citep{Caruana2015}, facilitate biomedical discoveries~\citep{Woo2017}, meet ethical and legal requirements~\citep{Amann2020,gdpr}, and ultimately facilitate doctor-AI communication and collaboration to leverage the strengths of both~\citep{2101.01524,Topol2019,Carter2017}. 

Applying XAI in clinical settings requires proper evaluation to ensure the explanation technique is both technically sound and clinically useful. Although existing works on XAI evaluation proposed many real-world evaluation objectives and metrics~\citep{Sokol2020,10.1145/3387166,VILONE202189,8400040,jacovi-goldberg-2020-towards,DBLP:journals/corr/abs-1806-08049,hase-bansal-2020-evaluating,doshivelez2017rigorous,DBLP:journals/corr/abs-1806-00069} (summarized in Supplementary Material S2 Table 1), there is not a canonical criterion on the goodness of explanation, and it is unknown which evaluation objectives are suitable for clinical applications. 
For the very limited emerging XAI evaluation works on medical image tasks, such as on retinal~\citep{10.1007/978-3-030-63419-3_3}, endoscopic~\citep{DESOUZA2021104578}, and chest X-Ray~\citep{Saporta2021.02.28.21252634, Arun2021} imaging tasks, the evaluation mainly focused on one criterion, which is how well the explanation agrees with clinical prior knowledge, without justification for the selection of such criterion and its clinical applicability. This evaluation criterion may be confounded by factors 
outside
XAI methods themselves, such as model training and spurious patterns in the data, as detailed in \S\ref{g4}. Furthermore, there are no clear guidelines on which evaluation objectives should be applied and prioritized to correspond to clinical requirements for AI explanation.

To answer Alex's questions and provide concrete support for the design and evaluation of clinical XAI, we propose the Clinical XAI Guidelines, which were developed with dual clinical and technical perspectives. The guidelines consist of five evaluation criteria:
The form of explanation is selected based on Guideline 1 (G1) Understandability and G2 Clinical relevance. 
The specific explanation technique for the selected form is chosen based on G3 Truthfulness, G4 Informative plausibility, and operational considerations on G5 Computational efficiency. Following the guidelines, we conducted a systematic evaluation of 16 commonly-used feature attribution map (heatmap) techniques on two multi-modal medical image tasks. We also formulated a novel and clinically pervasive problem of multi-modal medical image explanation, which is a generalized form of single-modal medical image explanation. We proposed the XAI evaluation metrics for this novel problem accordingly. The evaluation showed existing heatmap methods met G1, and partially met G2. But they did not meet G3 and G4, which suggests they are inadequate for clinical use.

Our key contributions are: 
\begin{enumerate}
\item We propose the Clinical XAI Guidelines grounded in both clinical and technical perspectives. The guidelines
support the selection and design of clinically viable XAI techniques for medical imaging tasks.
\item We conduct a systematic evaluation of multiple feature attribution map XAI algorithms on two medical imaging tasks to give a 
wholistic
evaluation of their adherence to the guidelines. 
\item Departing from the de-facto single modality explanation, we propose the clinically important but technically ignored problem of multi-modal medical image explanation and propose a novel metric: modality-specific feature importance (MSFI) to quantify and automate physicians' assessment of explanation plausibility.

\end{enumerate}

\paragraph{Roadmap} The manuscript is organized as follows: we first present the clinical XAI guidelines in \S\ref{gl}, with its key points highlighted in Table~\ref{table:gl} and Fig.~\ref{fig:gl}. We then present the systematic evaluation of 16 existing heatmap explanation methods based on the guidelines, with evaluation setup (\S\ref{prep}), evaluation methods (\S\ref{eval_method}), results (\S\ref{eval_result}), and discussions (\S\ref{discussion}).

\section{Clinical Explainable AI Guidelines }\label{gl}

By leveraging collective expertise in 
AI, clinical medicine, and human factor analysis, we developed the Clinical XAI Guidelines based on a thorough physician user study, our pilot XAI evaluation experiments~\citep{aaai2022,DBLP:journals/corr/abs-2107-05047}, and literature review (Supplementary Material S2 Table 1). The physician user study was conducted with 30 neurosurgeons on a glioma grading XAI prototype (Fig~\ref{fig:user_study}). We collected physicians' quantitative ratings on the heatmap explanation, and qualitative comments on the XAI system from the interview sessions and open-ended questionnaire. The qualitative data were used as the guidelines support from clinical aspect. The detailed user study findings and method are in Supplementary Material S1, and its related supporting sections were referred to in the paper starting with `U'.

Next, we present the Clinical XAI Guidelines, which is a checklist of five evaluation objectives to optimize a clinical XAI technique. They are categorized into three considerations: clinical usability, evaluation, and operation. For each objective in the guidelines, we list its key references from our user study or literature. 
The methods of assessment are also described to help identify if the objective is met. The guidelines and their key points are summarized in Table~\ref{table:gl}. The full version of the guidelines is in the Appendix.%

\begin{table*}
  \centering
  \begin{tabular}{L{1.7cm}L{5cm}L{4.5cm}L{4.5cm}}
    \toprule
    
    {\textit{Consideration}}
    & { \textit{Clinical XAI Guidelines }} 
    & { \textit{Ways of Assessment}}
    & { \textit{Key References}}\\
      
    \midrule
    \multirow{3}{*}{\makecell[tl]{\textit{Clinical}\\\textit{Usability}}}
    & \textbf{G1: Understandability} &&\\
        & Explanations should be easily understandable by clinical users without requiring technical knowledge.
        &  Sketch explanation forms 
        and show them to clinical users.
        &  \cite{jin2021euca}, 
        \cite{Sokol2020}; U3.3. Making AI transparent by providing information on performance, training dataset, and decision confidence.\\
        \cline{2-4}
            & \textbf{G2: Clinical relevance} &&\\
 & Explanation should be relevant to physicians' clinical decision-making pattern, and can support their clinical reasoning process.

        & Talk to or sketch prototypes with clinical users, to inspect if the explanation corresponds to their clinical reasoning process.

        &  U2.2. Resolving disagreement; U3. Clinical requirements of explainable AI.
\\
    \hline
    \multirow{3}{*}{\makecell[tl]{\textit{Evaluation}}} 
        & \textbf{G3: Truthfulness} &&\\

        &  
        Explanations should truthfully reflect the AI model decision process. This is the prerequisite for G4.

        &  Cumulative feature removal/addition test~\citep{DBLP:journals/corr/abs-2104-08782,NEURIPS2019_a7471fdc,DBLP:conf/nips/HookerEKK19,7552539,Lundberg2020,10.5555/3327757.3327875}; 
Synthetic dataset with known discriminative features as the ground truth~\citep{doshivelez2017rigorous,pmlr-v80-kim18d,DBLP:journals/corr/abs-1806-00069}. &        \cite{jacovi-goldberg-2020-towards,Sokol2020,DBLP:journals/corr/abs-2006-04948}; U2.3. Verifying AI decision, and calibrating trust.
\\

        \cline{2-4}
      & \textbf{G4: Informative plausibility}\\

        & 
        Users' judgment on explanation plausibility may inform users about AI decision quality, including potential flaws or biases.
        & 
        Statistical test on the correlation between AI decision quality measure and explanation plausibility measure \citep{adebayo2022post, Saporta2021.02.28.21252634}.  
        & 
       \cite{jacovi-goldberg-2020-towards}, \cite{Doshi-Velez2018};
U2. Clinical utility of explainable AI;
U5. Clinical assessment of explainable AI. \\

    \hline
    \multirow{2}{*}{\makecell[tl]{\textit{Operation}}} 
    & \textbf{G5: Computational efficiency} &&
    \\
            &  The speed to generate an explanation should be within clinical users' tolerable waiting time on the given task.
            & Understand how time sensitive the clinical task is, and record the speed and computational resources needed to generate an explanation.
           &\cite{Sokol2020}; U1.2.1. Decision support for time-sensitive cases, and hard cases. \\
    \bottomrule
  \end{tabular}
  \caption{The Clinical Explainable AI Guidelines for the design and evaluation of clinical explainable AI. Ways of assessment provide existing evaluation methods as references to assess if a guideline criterion is met. We list key references that supported the development of the guidelines. \\G - Guidelines, U - Physician user study findings (in Supplementary Material S1)
  }
  ~\label{table:gl}
\end{table*}

\subsection{Clinical usability considerations}
\noindent \textbf{Guideline 1: Understandability.}

The format and context of an explanation should be easily understandable by its clinical users. Users do not need to have technical knowledge in machine learning, AI, or programming to interpret the explanation.

\noindent \textbf{Guideline 2: Clinical relevance.}\label{g2}

The way physicians use explanations is to inspect the AI-based evidence provided by the explanation, and incorporate such evidence in their clinical reasoning process for downstream tasks, such as assessing the validity of AI decision, making a final decision on the case, improving their problem-solving skills, or making scientific discoveries (U2. Clinical utility of explainable AI; U1. Clinical utility of AI).
To make XAI clinically useful, the explanation information should be relevant to physicians' clinical decision-making pattern, and can support their clinical reasoning process.

For diagnostic/predictive tasks on medical images, a physician's image interpretation process includes two general steps: \textbf{1}) feature extraction: physicians first perform pattern recognition to localize key features and identify pathology of these features; \textbf{2}) reasoning on the extracted features: physicians perform medical reasoning and construct diagnostic hypotheses (differential diagnosis) based on the image feature evidence. A clinically relevant explanation should provide information corresponding to the above process, so that physicians can incorporate the explanation information into their medical image interpretation process (U3. Clinical requirements of explainable AI).

\subsection{Evaluation considerations}

\noindent \textbf{Guideline 3: Truthfulness.}

An explanation should truthfully reflect the model decision process. This is the fundamental requirement for a clinically oriented explanation, and an explanation method should fulfill the truthfulness requirement first prior to
G4: Informative plausibility.

\noindent \emph{Counterexample}: 

One of the main clinical utilities of explanation is that clinical users intuitively assess the plausibility of explanations (G4) to decide whether to take or reject the AI suggestion, and calibrate their trust in AI's current prediction on the case, or the AI model in general accordingly (U2.3). Users do so with an implicit assumption that explanations are the true representation of the model decision process. Violating truthfulness can lead to two significant consequences during physicians' use of explanation:

\textbf{1}. Clinical users may mistakenly reject AI's correct suggestion merely for the poor performance of the XAI method, which shows an unreasonable explanation.

\textbf{2}. If an XAI method is proposed or selected based on explanation plausibility objective only, rather than help clinical users to verify the decision quality, the explanation can be optimized to deceive clinical users with its seemingly plausible explanation, despite the wrong prediction from AI~\citep{DBLP:journals/corr/abs-2006-04948}.

\noindent \emph{Assessment method}:

The most common way to assess explanation truthfulness for feature attribution XAI methods in the literature is to gradually add or remove features from the most to the least important ones according to an explanation, and measure the model performance change~\citep{DBLP:journals/corr/abs-2104-08782,NEURIPS2019_a7471fdc,DBLP:conf/nips/HookerEKK19,7552539,Lundberg2020,10.5555/3327757.3327875, deyoung-etal-2020-eraser}. Another way is to construct synthetic evaluation datasets in which the ground-truth knowledge on the model decision process from input features to prediction is known and controlled~\citep{doshivelez2017rigorous,pmlr-v80-kim18d,DBLP:journals/corr/abs-1806-00069}.

\noindent \textbf{Guideline 4: Informative plausibility.}\label{g4}

The ultimate use of an explanation is to be interpreted and assessed by clinical users. Physicians intuitively use the assessment of explanation plausibility or reasonableness (i.e.: how reasonable the explanation is based on its agreement with human prior knowledge on the task) as a way to evaluate AI decision quality. This then allows multiple clinical utilities with XAI, including verifying AI's decisions (U2.3), calibrating trust in AI (U2.3), ensuring the safe use of AI, resolving disagreement with AI (U2.2), identifying potential biases, and making medical discoveries (U2.4). 
Informative plausibility assesses whether an XAI method can achieve its utility in helping users identify potential AI decision flaws and/or biases, i.e.: a plausible explanation for a right decision, and an implausible explanation for a wrong decision of AI. G3 Truthfulness is the gatekeeper of G4 Informative plausibility
to guarantee that the explanation truthfully represents the AI decision process.

\noindent \emph{Assessment method}:

To test whether explanation plausibility is informative to help users identify AI decision errors and biases, AI designers can assess the correlation between AI decision quality measures (such as model performance, calibrated prediction uncertainty, prediction correctness, and quantification of biased patterns) and plausibility measures~\citep{adebayo2022post, Saporta2021.02.28.21252634}. 

Since human assessment of explanation plausibility is usually subjective and susceptible to biases (U5.2. Bias and limitation of physicians’ quantitative rating), AI designers may consider quantifying the plausibility measure by abstracting the human assessment criteria into computational metrics for a given task. 
The quantification of human assessment is \emph{not} meant to directly select or optimize XAI methods for clinical use. Rather, XAI methods should be optimized for their truthfulness measures (G3). Quantifying plausibility is a means to validate the explanation's informativeness, i.e.: the effectiveness of XAI methods in their subsequent clinical utility to reveal AI decision flaws and/or biases, but not an XAI evaluation end goal in itself.
Quantifying plausibility can make such an informativeness validation process automatic, reproducible, standardizable, and computationally efficient. 
Similarly, the human annotation of important features according to physicians' prior knowledge, which is used to quantify plausibility, cannot be regarded as the ``ground truth'' of explanation, because explanations (given that they fulfill G3 Truthfulness) are still acceptable even if they are not aligned with human prior knowledge, but reveal the model decision quality or help humans identify new patterns and make biomedical discoveries.

\subsection{Operational consideration}
\noindent \textbf{Guideline 5: Computational efficiency}

Since many AI-assisted clinical tasks are time-sensitive decisions (U1.2.1. Decision support for time-sensitive cases, and hard cases), the selection or proposal of clinical XAI techniques needs to consider the computational time and resources. The wait time for an explanation should not be a bottleneck for the clinical task workflow.

\section{Evaluation problem setup}\label{prep}
In the previous section, we presented the Clinical XAI Guidelines. 
Next, we apply the guidelines to a specific problem on multi-modal medical image explanation. Multi-modal medical images, such as multi-parametric MRI, have indispensable diagnostic value in clinical settings. Nevertheless, their related explanation problem has not yet been explored in the technical community. We conduct a systematic evaluation on 16 commonly-used XAI methods to inspect whether their explanations on multi-modal medical images can fulfill the five objectives outlined in the Clinical XAI Guidelines and can be applied clinically.  
\subsection{Multi-modal medical imaging: clinical interpretation, learning, and explanation}
Our evaluation focuses on the novel problem of multi-modal medical image explanation. Multi-modal medical image explanation can be regarded as a generalized form of single-modal medical image explanation.
We present the clinical image interpretation process of multi-modal image, the clinical requirements for multi-modal image explanation, and different model learning paradigms on multi-modal medical image data.
\subsubsection{Multi-modal medical images and their clinical interpretation}
Multi-modal medical images consist of multiple image modalities or channels, where each modality captures a unique signal of the same underlying cells, tissues, lesions, or organs~\citep{MartBonmat2010}. Multi-modal images widely exist in the biomedical domain. For example,
different pulse sequences of magnetic resonance imaging (MRI) technique — T1 weighted, T2 weighted, or fluid-attenuated inversion recovery (FLAIR) modalities; dual-modality imaging of positron emission tomography-computed tomography (PET-CT)~\citep{pmid12072843}; CT images viewed at different levels and windows to observe different anatomical structures such as bones, lungs, and other soft tissues~\citep{HARRIS1993241}; 
multi-modal endoscopy imaging~\citep{Ray2017}; photographic, dermoscopic, and hyper-spectral images of a skin lesion~\citep{8333693,Zherebtsov2019}; multiple stained microscopic or histopathological images~\citep{Long2020, Song2013}. 

To interpret multi-modal images, doctors compare and combine modality-specific information to make diagnoses and differential diagnoses. For instance, in a radiology report on MRI, radiologists usually observe and describe \textit{anatomical} structures in T1 modality, and \textit{pathological} changes in T2 modality~\citep{cochard_netter_2012, Bitar2006}; doctors can infer the composition of a lesion (such as fat, hemorrhage, protein, fluid) by combining its signals from different MRI modalities~\citep{Patel2016}. In addition, some imaging modalities are particularly crucial for the diagnosis and management of certain diseases, such as a contrast-enhanced modality of CT or MRI for a suspected tumor case, and diffusion-weighted imaging (DWI) modality MRI for a suspected stroke case~\citep{Lansberg2000}. 

\subsubsection{Clinical requirements for multi-modal medical image explanation}\label{clin_req}
We summarize our findings on the clinical requirements for multi-modal medical image explanation based on our user study with neurosurgeons (U4 in Supplementary Material S1) on a glioma grading task with multi-modal brain MRI. 

To assess the plausibility of multi-modal explanation, physicians require the explanation to 1) prioritize the important image modality for the model's decision, and such prioritization may or may not necessarily need to be in concordance with physicians’ prior knowledge on modality prioritization; and 2) capture the modality-specific features. Such features may or may not be completely consistent with doctors’ prior knowledge, but should at least be
a subset and not \textcolor{black}{}deviate too much from clinical knowledge.

\subsubsection{Multi-modality learning}
There are three major paradigms to build convolutional neural network (CNN) models that learn from multi-modal medical images by fusing multi-modal features at the \textit{input}-level, \textit{feature}-level, or \textit{decision}-level~\citep{10.1007/978-3-030-32962-4_18}. 
Our evaluation covered two fusion settings at the \textit{input}-level (the brain tumor grading task) and \textit{feature}-level (the knee lesion identification task). For multi-modal fusion at the \textit{input}-level, the multi-modal images are stacked as input channels to feed a CNN. The modality-specific information is fused by summing up the weighted modality values in the first convolutional layer. For multi-modal image fusion at the \textit{feature}-level, each imaging modality is fed to its CNN branch individually to extract features first, and the image features are aggregated at a deeper layer.

\subsection{Clinical task, data, and model}
We include two clinical tasks in our evaluation on multi-modal medical image explanation: glioma grading on brain MRI, and knee lesion identification on knee MRI. Next, we describe the clinical task, medical imaging dataset, and the training of CNN models prepared for the evaluation.
\subsubsection{Glioma grading task}

\paragraph{Clinical task} As a type of primary brain tumors, gliomas are one of the most devastating cancers. Grading gliomas based on MRI provides physicians with indispensable information on a patient's treatment plan and prognosis. 
We focus on the task to classify gliomas into lower-grade (LGG) or high-grade gliomas (HGG).
\paragraph{Data} 
We used the publicly available BraTS 2020 dataset~\citep{Bakas2017}
and a BraTS-based synthetic dataset (described in \S\ref{syn}). Both are multi-modal 3D (BraTS) or 2D (synthetic) MRIs that consist of four modalities of T1, T1C (contrast enhancement), T2, and FLAIR. The BraTS dataset contains physician-annotated glioma localization masks that were used in the plausibility quantification.

\paragraph{Model}
For the BraTS dataset, we trained a VGG-like~\citep{DBLP:journals/corr/SimonyanZ14a} 3D CNN with six convolutional layers. It receives multi-modal 3D MRIs $X \in \mathbb{R}^{4 \times 240 \times 240 \times 155}$ of MRI modality, width, height, and depth respectively. 
We split the data into a training, validation, and test set with a 65\%, 15\%, 20\% split ratio. We trained five models using the same train/validation dataset and training scheme with different random seeds for model parameter initialization.
We used a weighted sampler to handle the imbalanced data. The models were trained with a learning rate $= 0.0005$, and batch size = 4. And training epoch was selected based on the accuracy on validation data.
The average accuracy on the test set for the five models is 89.46 $\pm$ 1.99\%.

For the synthetic glioma dataset, we fine-tuned a pre-trained DenseNet121 model~\citep{8099726} that receives 2D multi-modal MRI input slices of $X \in \mathbb{R}^{4 \times 256 \times 256}$ that represents MRI modality, width, and height.
We used the same training strategies as described above. The model achieves $95.70 \pm 0.06\%$ accuracy on the test set. 

\subsubsection{Knee lesion identification task}
\paragraph{Clinical task} MRI is the workhorse in diagnosing knee disorders with high accuracies~\citep{Rosas2009}. We focus on the task of identifying meniscus tear vs. intact based on knee MRI.
\paragraph{Data} We used the publicly available knee MRI dataset MRNet~\citep{Bien2018}. It consists of three modalities showing the knee structure from the coronal, sagittal, and axial view. The coronal view can be T1 weighted, or T2 weighted with fat saturation. The sagittal view is proton density (PD) weighted, or T2 weighted with fat saturation. Finally, the axial view is PD weighted with fat saturation.

We use bounding boxes of the meniscus as the representation of human prior knowledge in the explanation plausibility quantification. They were annotated by the first author who holds an M.D. degree based on knee MRI lesion interpretation principles~\citep{Rosas2009}. The bounding boxes are not exact annotations that localize the specific tear lesion, but only outline the anatomical location of the lateral and medial meniscus as a whole. This is meant to be closer to the practical real-world XAI evaluation scenario where only the least amount of annotation effort and domain expertise are required.

\paragraph{Model}
We used the same model architecture and training paradigm from the third place of MRNet challenge~\citep{Bien2018}, which fused multi-modal information at the feature level. We trained five models by only varying their random seeds for parameter initialization. The model performance area under the curve (AUC) on the validation set is $0.8395 \pm 0.0107$, which is equivalent to the reported ones in \cite{Bien2018}. The test AUC, however, is lower: $0.7934 \pm 0.0162$. 

\subsection{Post-hoc feature attribution explanation methods}
We chose feature attribution explanation methods based on user study assessment on G1 Understandability (detailed in Section \S\ref{g1}).
For feature attribution map methods, we focus on methods that are \textit{post-hoc}. This group of methods is a type of proxy models that probe the model parameters and/or input-output pairs of an already deployed or trained black-box model. In contrast, the \textit{ante-hoc} heatmap methods -- such as attention mechanism -- are predictive models with explanations baked into the training process.
We leave out the ante-hoc methods because such explanations are entangled in its specialized model architecture, which would introduce confounders in the evaluation.
We include 16 post-hoc XAI algorithms in our evaluation, which belong to two categories:
\begin{itemize}
    \item \textbf{Gradient-based}: Gradient~\citep{simonyan2014deep}, Guided BackProp~\citep{springenberg2015striving}, GradCAM~\citep{8237336}, Guided GradCAM~\citep{8237336}, DeepLift~\citep{10.5555/3305890.3306006}, Input$\times$Gradient~\citep{shrikumar2017just}, Integrated Gradients~\citep{10.5555/3305890.3306024}, Gradient Shap~\citep{NIPS2017_8a20a862},
Deconvolution~\citep{10.1007/978-3-319-10590-1_53}, Smooth Grad~\citep{smilkov2017smoothgrad}
    \item \textbf{Perturbation-based}: Occlusion~\citep{10.1007/978-3-319-10590-1_53,DBLP:conf/iclr/ZintgrafCAW17}, Feature Ablation, Shapley Value Sampling~\citep{CASTRO20091726}, Kernel Shap~\citep{NIPS2017_8a20a862}, Feature Permutation~\citep{JMLR:v20:18-760}, Lime~\citep{Ribeiro2016b} 

\end{itemize}
A detailed review of these algorithms and heatmap post-processing method are in Supplementary Material S2. 

\section{Evaluation method}\label{eval_method}

We present the systematic evaluation to inspect whether the commonly-used heatmap methods can be applied clinically to explain model decisions on multi-modal medical images. 
The evaluation follows the clinical XAI guidelines (\S\ref{gl}) to ensure the evaluation results can be an indicator for their suitableness in clinical settings. 
\subsection{Evaluating G1: Understandability}\label{g1}
We applied the end-user XAI prototyping method~\citep{jin2021euca} and asked our clinical collaborator to comment and select understandable explanation forms.
Based on the neurosurgeon's feedback and XAI technique availability, we targeted the explanation form of feature attribution map (namely, heatmap).

\subsection{Evaluating G2: Clinical relevance}
\begin{figure}[h]
    \centering
    \includegraphics[width=1\linewidth]{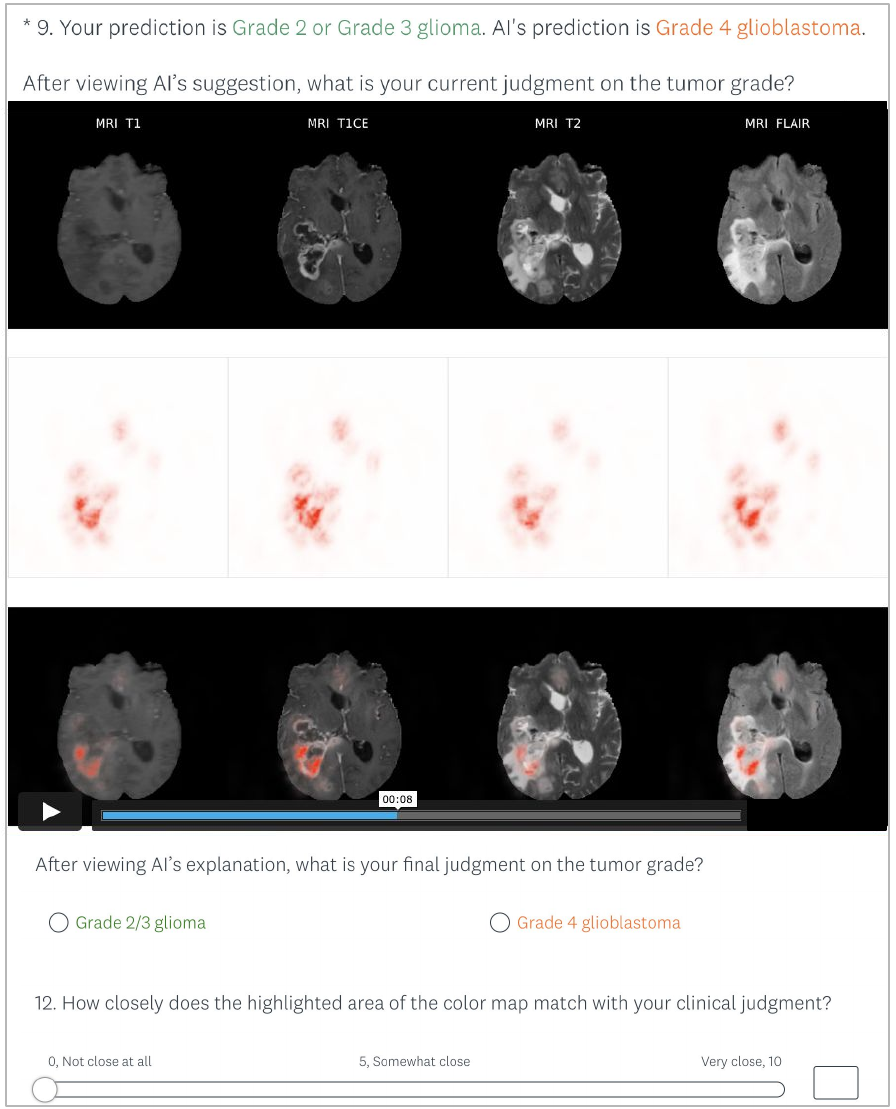}
    \caption{XAI prototype for the user study evaluation on G2 Clinical relevance. The low-fidelity XAI prototype is embedded in a survey: AI provides its prediction and heatmap explanation on a brain MRI, and the physician makes a decision assisted by AI suggestion and its explanation. In the embedded image, each column is an MRI modality. The first row shows the original MRI, the second row shows the heatmap explanation, and the third row shows the heatmap overlaid on MRI. Both MRI and heatmap are 3D images, and were presented as a video in the survey. The survey also collects physicians' ratings of the heatmap explanation.
    }
    \label{fig:user_study}
\end{figure}

To further identify the clinical relevance of heatmap explanation in the clinical usage scenario, we built an XAI prototype (Fig.~\ref{fig:user_study}) and conducted a user study with neurosurgeons. The user study method and findings are detailed in Supplementary Material S1.

\subsection{Evaluating G3: Truthfulness}
For the truthfulness assessment, we conducted cumulative feature removal and modality importance (MI) evaluation for the two clinical tasks, and proposed two novel metrics \textbf{$\text{$\Delta$AUPC}$} and \textbf{MI correlation} respectively. We also conducted a synthetic data experiment on the glioma grading task. 

\subsubsection{Cumulative feature removal}

\begin{figure}[!ht]
    \centering
    \includegraphics[width=1\linewidth]{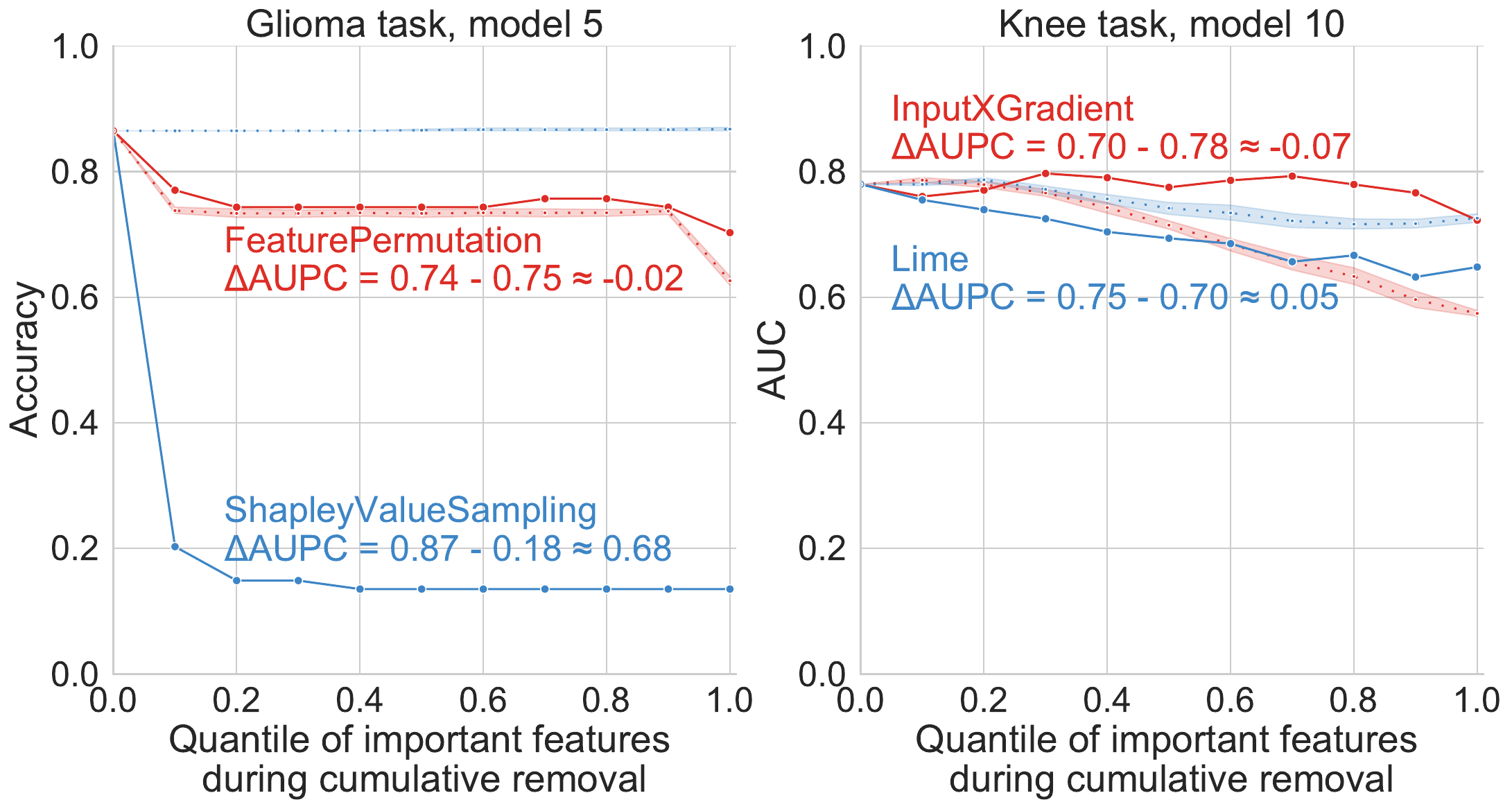}
    \caption{Feature perturbation curves for the cumulative feature removal experiment.
Feature perturbation curves in solid line are the model performance deterioration for an XAI method $\mathcal{H}$, and curves in dashed line are the XAI method counterpart baselines $\mathcal{H}_b$ of random feature removal. The random baseline experiment was repeated 15 times, thus the dashed line also has its 95\% confidence interval indicated as translucent error band. We show plots of the XAI method that has the highest (blue) and lowest $\Delta$AUPC score (red curve) from a model
    for both clinical tasks. $\text{AUPC}(\mathcal{H}_b)$ and $\text{AUPC}(\mathcal{H})$ which are used in the calculation of $\Delta\text{AUPC}$ are also indicated on the plot for each XAI method: $\Delta\text{AUPC} = \text{AUPC}(\mathcal{H}_b)-\text{AUPC}(\mathcal{H})$. Numbers reported in the subtraction are rounded to two decimal places.
    }
    \label{fig:acc_drop}
\end{figure}

To test if the heatmap highlighted regions are true important features to the model's decision, we cumulatively removed the input image features from the most to the least important ones according to the feature importance ranking quantile of an XAI algorithm $\mathcal{H}$. The removed features are replaced with a constant value (0 for glioma task, and modality mean for knee task). We then plotted a feature perturbation curve (PC) (Fig.~\ref{fig:acc_drop}) that shows the relationship of the cumulative feature removal to the model performance metric (accuracy for the glioma task, and AUC for the knee task). The area under PC ($\text{AUPC}(\mathcal{H})$) can be used to quantify the degree of performance deterioration during cumulative feature removal process: an XAI method $\mathcal{H}$ that indicates a more accurate feature importance ranking will lead to a faster performance deterioration, thus has a smaller AUPC. We proposed a new metric \textbf{$\text{$\Delta$AUPC}$} (difference of the area under the feature perturbation curve) defined as: $\text{$\Delta$AUPC}(\mathcal{H}) = \text{AUPC}(\mathcal{H}_b) -
\text{AUPC}(\mathcal{H})$, where $\text{AUPC}(\mathcal{H})$ and $\text{AUPC}(\mathcal{H}_b)$ are the area under the feature perturbation curve of an XAI method $\mathcal{H}$ and its corresponding baseline $\mathcal{H}_b$. $\text{$\Delta$AUPC}$ slightly modifies the above cumulative feature removal method in literature~\citep{DBLP:journals/corr/abs-2104-08782,NEURIPS2019_a7471fdc,DBLP:conf/nips/HookerEKK19,7552539,Lundberg2020,10.5555/3327757.3327875} by introducing a random baseline $\text{AUPC}(\mathcal{H}_b)$
for fair comparison among different XAI methods. For an XAI method $\mathcal{H}$, its corresponding random baseline $\mathcal{H}_b$ is generated by a random permutation of $\mathcal{H}$. For different XAI methods $\mathcal{H}$, the absolute numbers of highlighted image pixels/voxels are different, thus the performance deterioration measure may be confounded by the number of highlighted image regions.
$\text{$\Delta$AUPC}$ overcomes this to quantify the relative performance deterioration by comparing AUPC($\mathcal{H}$) with the AUPC of its corresponding random baseline $\mathcal{H}_b$. An XAI algorithm with a larger $\text{$\Delta$AUPC}$ indicates it can better identify important features for model prediction compared with its random baseline.

\subsubsection{Modality importance}\label{mi}
For multi-modal medical image explanation, we want to assess how truthfully a heatmap reflects the modality importance information used in the model decision process. This corresponds to the clinical requirements of modality prioritization (U4.2. The role and prioritization of multiple modalities). We first calculate the ground truth modality importance score using Shapley value method, then calculate the correlation between modality-wise sum of heatmap value and the ground truth as the modality importance correlation (\textbf{MI correlation}).

To determine the ground-truth modality importance, we use Shapley value from cooperative game theory~\citep{RM-670-PR}, due to its desirable properties such as efficiency, symmetry, linearity, and marginalism. In a set of $M$ modalities, Shapley value treats each modality $m$ as a player in a cooperative game play. It is the unique solution to fairly distribute the total contributions (in our case, the model performance) to each individual modality $m$.

We define the modality Shapley value $\varphi_m$ to be the ground truth modality importance score for a modality $m$. It is calculated as: 
\begin{align}\label{eq1}
    \varphi_m(v)\!=\!\sum_{c \subseteq \mathcal{M} \backslash\{m\}} \!\frac{|c| !(M-|c|-1) !}{M !}(v(c \cup\{m\})-v(c)),
\end{align}
where $v$ is the modality-specific performance metric (accuracy for the glioma task, and AUC for the knee task), and $\mathcal{M} \backslash\{m\}$ denotes all modality subsets $\mathcal{M}$ not including modality $m$. We constructed a modality subset $c$ by setting all values in a modality to 0 for modalities that were not included in the subset.

To measure the agreement of heatmaps' modality importance value with the ground truth modality Shapley value, for each heatmap, we define the estimated MI as the modality-wise sum of all positive values in the heatmap. MI correlation measures the MI ranking agreement between the ground-truth $\varphi$ and the estimated MI, calculated using Kendall's Tau-b ranking correlation.

\subsubsection{Synthetic data experiment}\label{syn}
The idea of constructing synthetic data to validate the truthfulness of an XAI method is that, we have the full control of the ground truth features that the model learned for its prediction, therefore, the ground truth features are also the ground truth for model decision rationale we want the explanation to capture. We can then assess the agreement between the explanation and the ground truth features using the same plausibility measure as detailed in \S\ref{method_plausibility}.

For multi-modal medical image tasks, according to the multi-modal medical image interpretation pattern identified in our user study (U4), we categorize the ground truth explanation information into: \textbf{1}. the relative importance of each modality to the prediction (i.e.: modality importance in \S\ref{mi}); and \textbf{2}. localization of the modality-specific features. We constructed a synthetic multi-modal brain MRI dataset on the glioma grading task with the two ground truth information corresponding to the prediction label. 

Specifically, to control the ground truth of feature localization, we use a GAN-based (generative adversarial network) tumor synthesis model developed by \cite{Kim2021} to generate two types of tumors and their segmentation masks, mimicking lower- and high-grade gliomas by varying their shapes (round vs. irregular~\citep{Cho2018}). 

To control the ground truth of modality importance, inspired by~\citep{pmlr-v80-kim18d}, 
we set tumor features on T1C modality to have 100\% alignment with the ground-truth label, and on FLAIR to have a probability of 70\% alignment, i.e., the tumor features on FLAIR correspond to the correct label with 70\% probability. The remaining modalities have 0 modality importance value, as they are designed to not contain class discriminative features. The model may learn to pay attention to either the less noisy T1C modality, or the more noisy FLAIR modality, or both. To determine their relative importance as the ground truth modality importance, 
we test the well-trained model on two test sets:

$\bullet$ TIC dataset: The dataset shows tumors only (without brain background) on all modalities. And the tumor shape has \textit{100\%} alignment with the ground-truth class label on \textit{T1C} modality, and 0\% alignment on FLAIR. Its test accuracy is denoted as $\text{Acc}_{\text{T1C}}$.
    
$\bullet$ FLAIR dataset: It has the same settings, but only differs in that the tumor shape has \textit{100\%} alignment with the ground-truth class label on \textit{FLAIR} modality, and 0\% alignment on T1C. Its test accuracy is denoted as $\text{Acc}_{\text{FLAIR}}$.

The test performance $\text{Acc}_{\text{T1C}}$ and $\text{Acc}_{\text{FLAIR}}$ indicate the degree of model reliance on that modality to make predictions. We use them as the ground truth modality importance. 
On the test set, $\text{Acc}_{\text{T1C}}=0.99$, $\text{Acc}_{\text{FLAIR }}=0$. In this way, we constructed a model with known ground truth of modality importance of 1 for T1C, and 0 for the remaining modalities. We then calculate the plausibility metric as the measure of truthfulness for the synthetic data.

\subsection{Evaluating G4: Informative plausibility}
Given an XAI method that meets G3: Truthfulness, to further validate whether clinical users can use their own assessment on explanation plausibility to judge decision quality and identify potential errors and biases, next we assess whether the human plausibility assessment is informative. We do so in two steps: \textbf{1}) proposing a novel plausibility metric -- modality-specific feature importance (\textbf{MSFI}) -- on multi-modal explanation task that 
bypasses physicians' manual assessment; and \textbf{2}) testing the correlation between plausibility metric and decision quality metric.

\subsubsection{Quantifying plausibility}\label{method_plausibility}
\begin{figure*}[h]
    \centering
    \includegraphics[width=1\linewidth]{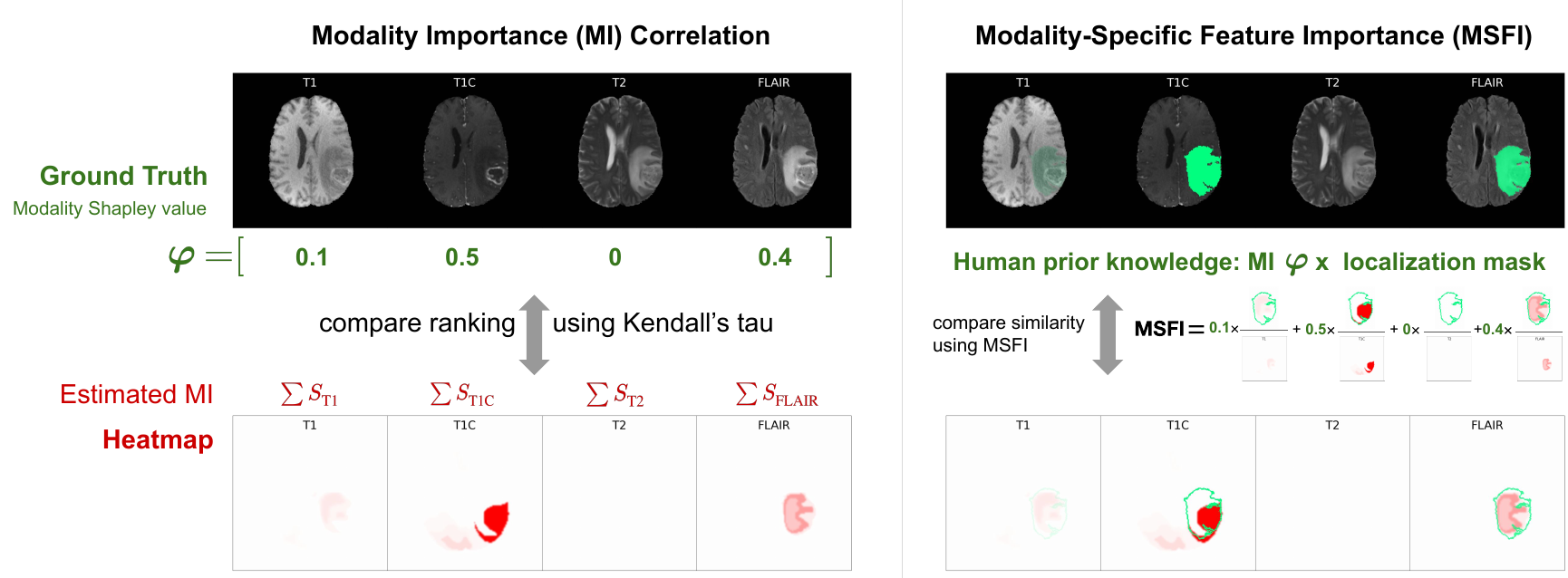}
    \caption{Illustration of the novel modality importance correlation and MSFI metrics on multi-modal medical image explanation.}
    \label{fig:eval_outline}
\end{figure*}
To quantify how reasonable the explanation is to human judgment and facilitate subsequent validation of using such plausibility information for AI decision verification, we used an existing metric feature portion (\textbf{FP}), and proposed a novel metric modality-specific feature importance (MSFI) designed for multi-modal medical image explanation based on its clinical requirements (\S\ref{clin_req}). Both metrics quantify the agreement of heatmap highlighted regions with human prior knowledge.

FP assesses, among the highlighted regions in the heatmap, how many of them agree with human prior knowledge. It is calculated as:
\begin{align}
\text{FP} &= \frac{ \sum_i \mathbbm{1}
( L^i >0 ) \odot S^i }{ \sum_i 
S^i  }
\end{align}
where $S$ is a heatmap, with $i$ denoting the spatial location. $L$ is the human-annotated feature masks, with $L^i>0$ outlining the spatial location of the feature. $\mathbbm{1}$ is the indicator function that selects the heatmap values inside the feature mask. 

To abstract the clinical requirements for multi-modal medical image explanation (U4. Multi-modal medical image interpretation and clinical requirements for its explanation), we propose a novel plausibility metric MSFI for multi-modal explanation (Fig.~\ref{fig:eval_outline}). It combines the assessment of feature localization with modality prioritization, by multiplying FP with modality importance value  modality-wise.
Specifically, MSFI is the portion of heatmap values $S_m$ inside the feature localization mask $L_m$ for each modality $m$, weighted by MI $\varphi_m$ which is normalized to $[0, 1]$ to have a comparable range with FP.

\begin{align}
\widehat{\text{MSFI}} &=\sum_{m}  \varphi_m \frac{ \sum_i \mathbbm{1}
( L_m^i >0 ) \odot S_m^i }{ \sum_i 
S_m^i  },\\
\text{MSFI} &= \frac{\widehat{\text{MSFI}}} {\sum_{m} \varphi_m},
\end{align}
where $\widehat{\text{MSFI}}$ is unnormalized, and ${\text{MSFI}}$ is the normalized metric in $[0, 1]$. 
A higher MSFI score 
indicates a heatmap is more agreeable with clinical prior knowledge regarding capturing the important modalities and their localized features. 
MSFI can be regarded as a general form of FP that generalizes the feature portion calculation from single-modality to multi-modality images. 

Instead of asking physicians to manually assess plausibility for a few explanations (the questionnaire in Fig.~\ref{fig:user_study} demonstrates such process), whose rating may be susceptible to cognitive biases (U5.2. Bias and limitation of physicians’ quantitative rating), quantifying plausibility bypasses humans' manual assessment, standardizes and automates the assessment process, and can assess multiple XAI methods using one set of annotated data. 

In addition, although plausibility quantification requires annotations to represent human prior knowledge, the human prior knowledge annotation may not necessarily need to be as exact as feature segmentation masks, because MSFI and FP only penalize for regions outside the annotation mask\footnote{In comparison, we did not use the intersection over union (IoU) metric commonly used in computer vision, because compared to MSFI or FP that penalizes only for false positives, IoU also penalizes for false negatives, which require the annotations to be exact.} $L$. 
Therefore, the annotation can be in the form of segmentation masks, bounding boxes, or landmarks. In our evaluation, we used tumor segmentation masks for the glioma task, and bounding boxes for the knee task. The annotations may not even need to be annotated by humans. It can be generated by training an AI model on a few annotated data points, or using trained models on feature segmentation/localization tasks. 

\subsubsection{Testing for plausibility informativeness}
The indispensable step after plausibility quantification is to validate the clinical utility of using explanations to verify AI decision quality. 
We measure AI decision quality by using 1) the soft output probability, and 2) the hard thresholding model prediction correctness on the two classification tasks.
We then test the correlation between prediction probability and plausibility, and test for identically distributed plausibility for different prediction correctness groups. Unless otherwise stated, we use a significance level $\alpha = 0.05$ for two-sided statistical test.

\subsection{Evaluating G5: Computational efficiency}
We recorded the computational time to generate each heatmap on a computer with 1 GTX Quadro 24 GB GPU and 8 CPU cores, and on a computing cluster with similar hardware configurations.

\section{Evaluation result}\label{eval_result}

We report evaluation results on whether the commonly-used 16 heatmap methods are clinically feasible by fulfilling the guidelines on the two clinical tasks with multi-modal medical images. All results were reported on the test dataset.

\subsection{Evaluating G1 Understandability and G2 Clinical relevance}\label{g1g2}

In our user study, although physicians did not express difficulty in understanding the meaning of heatmap as important regions for AI prediction (G1: Understandability is met), 
the heatmap explanation is not completely clinically relevant, as physicians were perplexed by the highlighted areas regardless of whether these areas align with their prior knowledge or not. This may be due to heatmap explanation only performing half of the clinical image interpretation step of feature localization, it lacks pathological description of important features, let alone to perform reasoning on these features (U3.1. Limitations of existing heatmap explanation). Therefore, the heatmap explanation only partially fulfills G2 Clinical relevance. 

\begin{table*}[]

    \centering
    \begin{tabular}{m{0.2\linewidth} *{2}{>{\centering\arraybackslash}m{0.12\linewidth}}|
    *{2}{>{\centering\arraybackslash}m{0.12\linewidth}}|
    *{1}{>{\centering\arraybackslash}m{0.12\linewidth}}
    }
    \toprule

    & \multicolumn{2}{c}{Cumulative  feature  removal} & \multicolumn{2}{c}{Modality importance correlation} & Synthetic data experiment\\
    & \multicolumn{2}{c}{\textbf{$\Delta$AUPC} \small [-1, 1]} & \multicolumn{2}{c}{\textbf{MI correlation} \small [-1, 1]} & \textbf{MSFI} \small [0, 1]\\

    \cline{2-6}
     & Glioma & Knee & Glioma & Knee  & Synthetic glioma\\

\hline
Deconvolution
 &0.38$\pm$0.14 & -0.04$\pm$0.04  & 0.46$\pm$0.28  & -0.47$\pm$0.51 &   0.04$\pm$0.02  \\
\hline
DeepLift
& 0.16$\pm$0.10  &NaN&  0.60$\pm$0.33   & NaN & 0.22$\pm$0.23   \\
\hline
Feature Ablation
& 0.34$\pm$0.11 &  -0.02$\pm$0.04 &  0.60$\pm$0.43   & 0.05$\pm$0.64  & 0.19$\pm$0.23   \\
\hline
Feature Permutation
& -0.03$\pm$0.08   &NaN& NaN &  NaN & 0.08$\pm$0.07 \\
\hline
GradCAM
& 0.22$\pm$0.16 & NaN& NaN & NaN & 0.02$\pm$0.02   \\
\hline
Gradient
& 0.09$\pm$0.02 & -0.05$\pm$0.02 & 0.49$\pm$0.41    & -0.52$\pm$0.51 & 0.19$\pm$0.13  \\ 
\hline
Gradient Shap
& 0.18$\pm$0.12 & -0.02$\pm$0.03 & \textbf{0.64$\pm$0.31}   & -0.29$\pm$0.54 & 0.22$\pm$0.19   \\
\hline
Guided BackProp
& \textbf{0.53$\pm$0.09} &  -0.04$\pm$0.03  & 0.57$\pm$0.21   & -0.44$\pm$0.53 & $*$\textbf{0.49$\pm$0.21}  \\
\hline
Guided GradCAM
& \textbf{0.53$\pm$0.09} &  NaN& 0.56$\pm$0.23   &NaN & \textbf{0.42$\pm$0.29}   \\
\hline
Input$\times$Gradient
& 0.16$\pm$0.11 & -0.05$\pm$0.03  & \textbf{0.64$\pm$0.29}  & -0.35$\pm$0.55  & \textbf{0.23$\pm$0.14}   \\
\hline
Integrated Gradients
&0.18$\pm$0.12 & -0.04$\pm$0.02  & 0.63$\pm$0.31   &  0.24$\pm$0.64  & 0.22$\pm$0.19   \\
\hline
Kernel Shap
& 0.31$\pm$0.10  & \textbf{0.00$\pm$0.03} &  NaN &  $*$\textbf{0.33$\pm$0.58} &  0.08$\pm$0.08   \\
\hline
Lime
& \textbf{0.51$\pm$0.08}  & \textbf{0.00$\pm$0.04 } & 0.57$\pm$0.42    & $*$\textbf{0.35$\pm$0.58} & 0.05$\pm$0.07  \\
\hline
Occlusion
& 0.21$\pm$0.08 & -0.01$\pm$0.02  & 0.58$\pm$0.45  & -0.32$\pm$0.54& 0.22$\pm$0.25   \\
\hline
Shapley Value Sampling
& \textbf{0.51$\pm$0.10}  & \textbf{0.00$\pm$0.04 } & 0.59$\pm$0.37   & $*$\textbf{0.35$\pm$0.50} & 0.10$\pm$0.10   \\
\hline
Smooth Grad
& 0.48$\pm$0.08 & -0.05$\pm$0.03 & $*$\textbf{0.72$\pm$0.24}  & -0.43$\pm$0.57  & 0.03$\pm$0.02  \\
\bottomrule
    \end{tabular}
    \caption{\textbf{Evaluation results on Guideline 3 - Truthfulness}. 
    The table shows mean $\pm$ std for each XAI algorithm on three evaluation metrics: $\Delta$AUPC, MI correlation, and MSFI on the synthetic data. Metrics have their range indicated. For all metrics, a higher value is better. Top three results on a metric are in bold, with a $*$ indicating the XAI algorithm performed significantly better than others. ``NaN'' in the glioma task is because the heatmap is not modality-specific and the correlation is not computable. ``NaN'' in the knee task is because the XAI method was not included in the evaluation. XAI methods are in alphabetic order. 
    }
    \label{tab:eval_result_g3}
\end{table*}

\subsection{Evaluating G3: Truthfulness}

The evaluation results on G3 Truthfulness of all three evaluation experiments are shown in Table~\ref{tab:eval_result_g3}. The $\Delta$AUPC metric on cumulative feature removal experiment is a global metric that runs on the whole test set, and we reported the metric mean $\pm$ standard deviation (std) of five models on the same test set, and used it to compare the XAI method performances; whereas the other evaluation metrics are local and run on individual data point, and we reported their mean $\pm$ std of five models by aggregating all test data points, and conducted Friedman and post-hoc Nemenyi test to identify the top ranking XAI methods. Using Kendall's Tau-b ranking correlation, we also tested the performance ranking (using the mean of a metric) correlation between the glioma and knee tasks, to see if the performance on one task can be generalized to another task. 

For the cumulative feature removal experiment that examines the fine-grained \textit{feature}-level explanation truthfulness of XAI methods to the model decision process, the performances of the examined XAI methods on glioma and knee tasks differ a lot: on the glioma task, Guided BackProp, Guided GradCAM, Lime, Shapley Value Sampling, and Smooth Grad were the top-ranked algorithms with an average $\Delta$AUPC around 0.5, and their performances were relatively stable across different models.
Whereas on the knee task, all XAI methods performed poorly with their $\Delta$AUPC scores around 0, which indicates the examined XAI methods did not differ from the baseline of random heatmaps. 
In addition, when comparing the glioma and knee tasks on the XAI method rankings based on mean $\Delta$AUPC, there was not a statistically significant correlation using Kendall's Tau-b ($\tau_b = 0.24, p=0.31$), indicating the performance of 
XAI methods may only be specific to a task and not generalizable.

For the MI correlation experiment that examines the coarse-grained \textit{modality}-level explanation truthfulness of XAI methods to the model decision process, on the glioma task, the importance ranking of heatmaps modalities showed weak to moderate positive correlations with the ground-truth modality Shapley values. Among the examined 13 XAI methods, there was a statistically significant difference of mean MI correlation using Friedman test, $\chi^2(12) = 223.3, p<0.001$. A post-hoc Nemenyi test showed only Smooth Grad had a statistical significance higher performance than the rest of XAI methods ($p<0.01$). On the knee task, the examined 12 XAI methods showed from moderate negative to weak positive correlations with the ground-truth Shapley values, and there was a statistically significant difference of mean MI correlation using Friedman test, $\chi^2(11) = 912.6, p<0.001$. A post-hoc Nemenyi test showed Lime, Shapley Value Sampling, and Kernel Shap had a statistical significance higher performance than the rest of XAI methods ($p<0.01$). 
Furthermore, the MI correlation performance ranking on one task did not migrate to another, with a statistically insignificant Kendall's Tau-b ranking correlation test, $\tau_b = 0.13, p=0.65$.

For the synthetic data experiment on the glioma task that examines both \textit{modality}- and \textit{feature}-level truthfulness of XAI methods to the model decision process, the MSFI scores were generally in the low range, and no XAI method achieved an average MSFI score above 0.5. Among these, only Guided BackProp outperformed other XAI methods with statistical significance ($p<0.01$) using a post-hoc Nemenyi test after a significant Friedman test ($\chi^2(15) = 1540.6, p<0.001$). Since the synthetic data evaluation combined both the coarse-grained modality-level (MI correlation) and the fine-grained feature-level explanation truthfulness ($\Delta$AUPC), we further tested whether the XAI method performance on the synthetic data can be used to guide the selection of XAI on the original real-patient data on glioma task. Kendall's Tau-b correlation test showed that the MSFI mean score ranking of the synthetic data experiment had no statistically significant ranking correlation with MI correlation ($\tau_b = 0.08, p=0.77$), and with $\Delta$AUPC ($\tau_b = -0.05, p=0.82$).

In summary, on the glioma task, the only XAI methods that outperformed others on feature-level ($\Delta$AUPC and MSFI on synthetic data experiment) and modality-level (MI correlation) explanation truthfulness evaluations are Guided BackProp. Despite this, the performances of the top XAI methods were  around 0.5 compared to the ground truth or out-performed the random baseline. Since there is no benchmark, and the relative weights for individual evaluation metrics are unknown, the fulfillment of G3 Truthfulness may be dependent on the specific task and its clinical importance. 
On the knee task, all the examined XAI methods failed to meet G3 Truthfulness due to their low evaluation performances on both modality- and feature-level truthfulness. In addition, the good-performing XAI method on one clinical task did not generalize to another task.

\subsection{Evaluating G4: Informative plausibility}

\begin{table*}[!h]

    \centering
    \resizebox{\textwidth}{!}{\begin{tabular}{m{0.2\linewidth} *{2}{>{\centering\arraybackslash}m{0.1\linewidth}}|*{3}{>{\centering\arraybackslash}m{0.1\linewidth}}|
    *{3}{>{\centering\arraybackslash}m{0.1\linewidth}}
    }
    \toprule
     & \multicolumn{2}{c}{MSFI correlation w/ pred. prob.} &  \multicolumn{3}{c}{Testing for plausibility informativeness - Glioma} & \multicolumn{3}{c}{Testing for plausibility informativeness - Knee}\\

    \cline{2-9}
    & Glioma & Knee & Stat. Sig. & Right Pred. & Wrong Pred. & Stat. Sig. & Right Pred. & Wrong Pred.\\

\hline

Deconvolution & 0.41 $^*$ & -0.08 $^*$  &   NS & 0.47 (0.44,0.49) & 0.55 (0.44,0.57) & NS & 0.23 (0.22,0.24) & 0.23 (0.22,0.24)\\ \hline

DeepLift & 0.49 $^*$ &  NaN &   $\star$ & 0.83 (0.79,0.85) & 0.75 (0.55,0.81) & NaN & NaN & NaN\\ \hline

FeatureAblation & \textbf{0.59} $^*$ & -0.16 $^*$  &   $\star\star$ & 0.70 (0.66,0.75) & 0.48 (0.29,0.67) & NS & 0.16 (0.15,0.16) & 0.18 (0.17,0.19)\\ \hline

FeaturePermutation & 0.19 $^*$ &  NaN &   NS & 0.29 (0.22,0.35) & 0.21 (0.06,0.32) & NaN & NaN & NaN\\ \hline

GradCAM & 0.18 $^*$ &  NaN &   NS & 0.04 (0.04,0.05) & 0.04 (0.02,0.05) & NaN & NaN & NaN\\ \hline

Gradient & 0.41 $^*$ & -0.08 $^*$  &   NS & 0.49 (0.46,0.51) & 0.50 (0.33,0.54) & NS & 0.24 (0.24,0.25) & 0.24 (0.24,0.25)\\ \hline

GradientShap & 0.49 $^*$ & -0.09 $^*$  &   $\star$ & 0.78 (0.75,0.80) & 0.70 (0.52,0.76) & NS & 0.23 (0.22,0.23) & 0.24 (0.22,0.24)\\ \hline

GuidedBackProp & 0.41 $^*$ & -0.07 &   NS & 0.78 (0.74,0.79) & 0.76 (0.57,0.82) & NS & 0.25 (0.25,0.26) & 0.25 (0.25,0.27)\\ \hline

GuidedGradCAM & 0.37 $^*$ &  NaN &   NS & 0.82 (0.80,0.85) & 0.80 (0.54,0.86) & NaN & NaN & NaN\\ \hline

Input$\times$Gradient & \textbf{0.57} $^*$ & -0.08 $^*$  &   $\star$ & 0.77 (0.75,0.79) & 0.69 (0.46,0.76) & NS & 0.23 (0.23,0.24) & 0.24 (0.24,0.25)\\ \hline

IntegratedGradients & 0.50 $^*$ & -0.08 &   $\star$ & 0.78 (0.75,0.82) & 0.71 (0.51,0.76) & NS & 0.22 (0.21,0.23) & 0.23 (0.22,0.23)\\ \hline

KernelShap & 0.36 $^*$ & -0.13 $^*$  &   NS & 0.20 (0.16,0.23) & 0.15 (0.03,0.35) & NS & 0.15 (0.15,0.16) & 0.17 (0.16,0.18)\\ \hline

Lime & 0.36 $^*$ & -0.13 $^*$  &   NS & 0.23 (0.19,0.27) & 0.19 (0.12,0.25) & NS & 0.16 (0.16,0.17) & 0.18 (0.17,0.18)\\ \hline

Occlusion & \textbf{0.60} $^*$ & -0.07 &   $\star\star\star$ & 0.55 (0.54,0.58) & 0.24 (0.14,0.44) & NS & 0.20 (0.19,0.21) & 0.20 (0.19,0.21)\\ \hline

ShapleyValueSampling & 0.53 $^*$ & -0.10 $^*$  &   NS & 0.58 (0.54,0.61) & 0.38 (0.27,0.59) & NS & 0.17 (0.16,0.17) & 0.18 (0.17,0.19)\\ \hline

SmoothGrad & 0.36 $^*$ & -0.03 &   NS & 0.39 (0.37,0.40) & 0.39 (0.32,0.42) & NS & 0.24 (0.23,0.24) & 0.24 (0.23,0.25)\\

\bottomrule
    \end{tabular}}
    \caption{\textbf{Evaluation results on Guideline 4 - Testing for plausibility informativeness}. In the column: MSFI correlation with prediction probability, the statistically significant Spearman's correlations are marked with $*$, and bold text highlights the top three positively correlated XAI methods. In the column: Testing for plausibility informativeness on glioma and knee task, we report the significant level and MSFI score (median and 95\% confidence interval) of right and wrong predictions.
    The statistical significance are from the upper-tailed Mann–Whitney U test: $\star$ indicates $p<0.025$; $\star\star$ for $p<0.005$; $\star\star\star$ for
    $p<0.0005$; NS for not significant. 
    ``NaN'' in the knee task is because the XAI method was not included in the evaluation. XAI methods are in alphabetic order. 
    }
    \label{tab:eval_result_g4}
\end{table*}

\begin{figure*}[!th]
    \centering
    \includegraphics[width=1\textwidth]{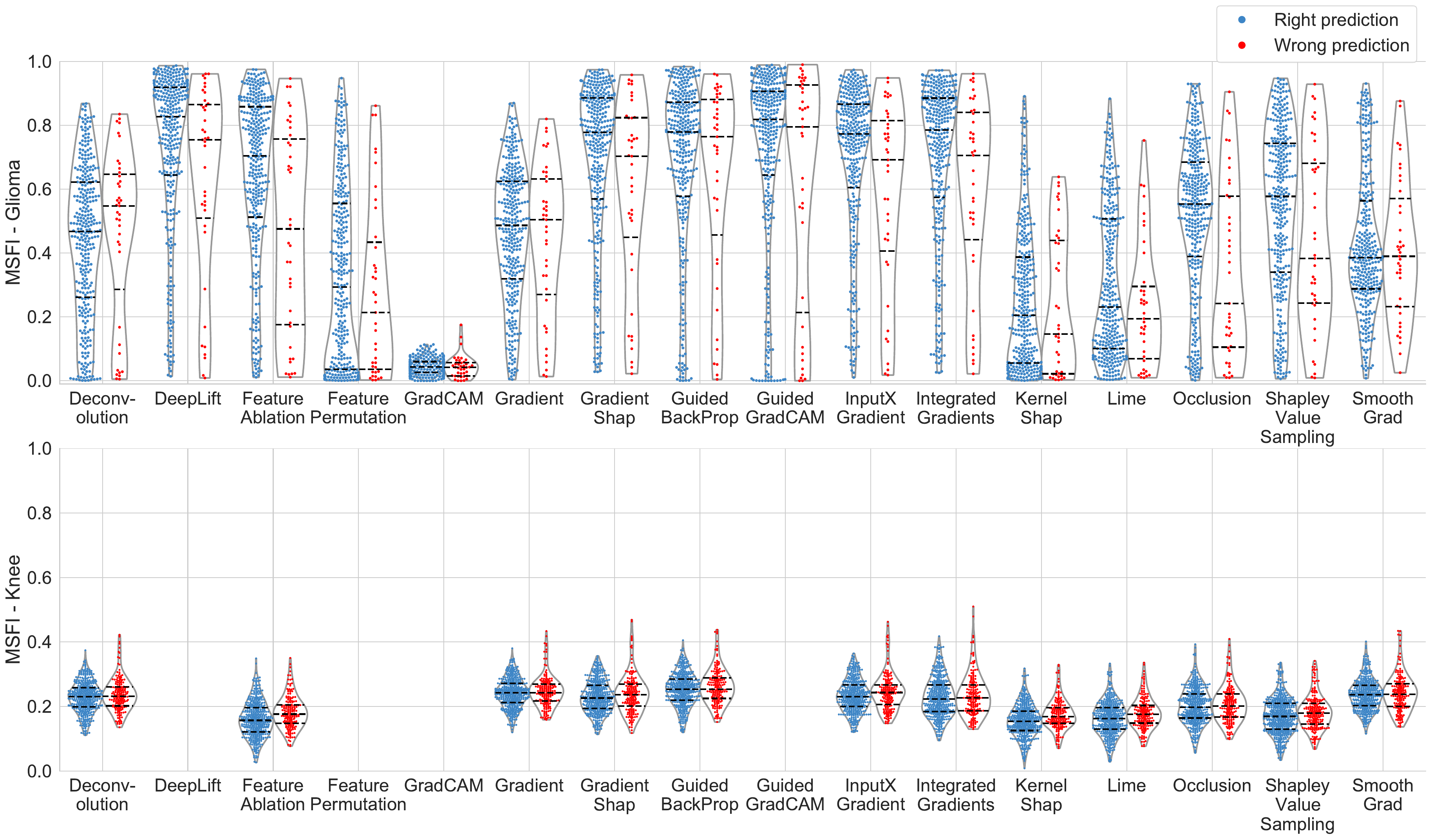}
    \caption{\textbf{Evaluation results on Guideline 4 - Testing for plausibility informativeness.} 
    For each heatmap method ($X$-axis), the violin and swarm plots show the plausibility quantification score distribution of MSFI for the right (blue, left) and wrong (red, right) predictions on the glioma (top) and knee task (bottom). Each dot is a data sample in the test set, and we aggregate results from five similarly-trained models. $Y$-axis is the MSFI measure, with a higher score indicating more agreeable of a heatmap with clinical prior knowledge on modality prioritization and feature localization. The black dashed lines indicate the quartiles of each distribution.
    }
    \label{fig:g4}
\end{figure*}

\subsubsection{Quantifying plausibility}

Physicians' average quantitative rating on heatmap quality had a higher Pearson's r correlation with MSFI ($r(53)=0.59$, $p<0.001$) compared with FP ($r(53)=0.57$, $p<0.001$). 
Therefore, we resorted to quantifying the human assessment of explanation plausibility using MSFI score, while reporting the results using FP measure in Supplementary S2.
In addition, physicians' inter-rater agreement
on the heatmap quality assessment was low: Krippendorff's Alpha is 0.528 (cutoff value $\geq$ 0.667~\citep{krippendorff2004content}), and Fleiss' kappa is 0.009 (with 1 for perfect agreement and 0 for poor agreement). This indicates that doctors' judgment of heatmap quality could be very subjective, which aligns with qualitative findings on U5.2. Bias and limitation of physicians’ quantitative rating.

\subsubsection{Testing for plausibility informativeness}\label{testing_for_informativeness}
Since G3 Truthfulness is the prerequisite for G4 on plausibility informativeness, 
it is less meaningful to conduct 
plausibility informativeness assessment for XAI methods that did not fulfill G3 Truthfulness. Nevertheless, we reported the full evaluation results for all XAI methods as a reference.

To examine the correlation between plausibility measure MSFI and model prediction probability, we computed their non-parametric Spearman correlation (Table~\ref{tab:eval_result_g4}).
For the glioma task, the plausibility measure MSFI of all XAI methods had a weak to moderate positive correlation with the model prediction probability, and the correlations were all statistically significant ($p<0.001$). Occlusion, Feature Ablation, and Input$\times$Gradient were the top three highly correlated XAI methods. For the knee task, all methods had a negative weak correlation with model prediction probability that may or may not show statistical significance.

The above model output probability may not be well calibrated~\citep{DBLP:journals/corr/GuoPSW17}, thus may not be a good indicator for model decision quality.
We then resorted to model prediction correctness as the definitive indicator for decision quality. Using the non-parametric Mann-Whitney U test~\citep{10.2307/2236101}, we tested the upper-tailed alternative hypothesis that the distribution of MSFI on the correctly predicted data group is significantly higher than the incorrectly predicted one. The resulting significance level for each XAI algorithm is shown in Table~\ref{tab:eval_result_g4}. 
For some XAI methods such as Occlusion and Feature Ablation, despite they showed statistically higher MSFI scores on the right prediction data group compared to the wrong prediction one, by further inspecting their distributions (Fig.~\ref{fig:g4}-top), the ranges of correctly and incorrectly predicted data points largely overlapped with each other. This may hinder the application of XAI methods for clinical users to identify potential decision flaws based on their plausibility judgment of the explanation, because the right and wrong predictions could have the same range of MSFI scores. For the knee task, all XAI methods failed to reject the null hypothesis, with the right and wrong prediction data points having similar MSFI score distributions (Fig.~\ref{fig:g4}-bottom).
Similar to the evaluation on G3, in G4 evaluation, the examined XAI methods did not exhibit the same performance pattern on the glioma and knee task.

The testing for plausibility informativeness on glioma task showed that, despite the overall range of the correctly and incorrectly predicted data points overlapping with each other, for some XAI methods, the Mann-Whitney U test still showed statistically higher MSFI for the correctly predicted data points than the incorrectly predicted ones. Further analysis showed that the statistical test result was confounded by different MSFI distributions on the two classes of LGG and HGG: for all XAI methods, both the predicted and ground-truth HGG class had a significantly higher ($p<0.0005$) MSFI score compared to the predicted or ground-truth LGG class. The different distributions of MSFI on LGG and HGG classes influenced the results on testing for informative plausibility. To remove the influence of this confounder, we then conducted testing for plausibility informativeness \textit{conditioned} on each class, and it yielded similar results as the above unconditioned one: 
when conditioned on HGG prediction, only Occlusion and Feature Ablation showed significantly higher MSFI for the rightly predicted data compared to the wrongly predicted ones, with $p = 0.003$ and $0.01$ respectively. None of the XAI methods showed statistical significance when conditioned on LGG prediction. The visualization of MSFI conditioned on either HGG or LGG prediction, however, still showed range overlapping for the right and wrong predictions (Supplementary S2 Fig. 16). This indicates the examined XAI methods, both the unconditioned one and the one conditioned on each predicted class, failed the testing for informative plausibility.
The same analysis on the knee task did not show statistically different MSFI on right and wrong predictions conditioned on each predicted class.
The above analysis is detailed in Supplementary S2 \S4.3.2.

Based on the results on testing for plausibility informativeness, the examined XAI methods did not meet G4 Informative plausibility neither on the glioma nor on the knee task.

\subsection{Evaluating G5: Computational efficiency}
\begin{table}[]

    \centering
    \begin{tabular}{m{0.3\linewidth} *{3}{>{\centering\arraybackslash}m{0.19\linewidth}}
    }
    \toprule
    & \multicolumn{3}{c}{Computational time} \\
    & \multicolumn{3}{c}{\textbf{seconds}} \\

    \cline{2-4}
     & Glioma & Synthetic Glioma & Knee\\
\hline
Deconvolution & 2.1$\pm$1.2 &  1.3$\pm$0.0 &  2.6$\pm$2.1  
\\ \hline
DeepLift & 4.6$\pm$2.0 &  2.2$\pm$0.0 &  NaN  
\\ \hline
FeatureAblation & 82$\pm$25 &  58$\pm$1.5 &  98$\pm$102  
\\ \hline
FeaturePermutation & 10.1$\pm$2.1 &  15.2$\pm$0.4 &  NaN  
\\ \hline
GradCAM & 0.7$\pm$0.3 &  0.3$\pm$0.0 &  NaN  
\\ \hline
Gradient & 2.2$\pm$1.3 &  1.1$\pm$0.0 &  2.6$\pm$2.2  
\\ \hline
GradientShap & 7.8$\pm$3.3 &  5.0$\pm$0.1 &  2.8$\pm$2.2  
\\ \hline
GuidedBackProp & 2.1$\pm$1.2 &  0.9$\pm$0.0 &  2.3$\pm$1.7  
\\ \hline
GuidedGradCAM & 2.8$\pm$1.5 &  1.2$\pm$0.0 &  NaN  
\\ \hline
Input$\times$Gradient & 2.1$\pm$1.2 &  1.1$\pm$0.0 &  2.6$\pm$2.2  
\\ \hline
IntegratedGradients & 67$\pm$34 &  49$\pm$0.9 &  113$\pm$79  
\\ \hline
KernelShap & 243$\pm$87 &  93$\pm$1.6 &  382$\pm$388 
\\ \hline
Lime & 449$\pm$141 &  154$\pm$2.6 &  507$\pm$523  
\\ \hline
Occlusion & 1713$\pm$21 &  27$\pm$3.5 &  672$\pm$255  
\\ \hline
ShapleyValueSampling & 2205$\pm$693 &  1595$\pm$228 &  1990$\pm$2021  
\\ \hline
SmoothGrad & 14.4$\pm$6.8 &  9.5$\pm$0.1 &  24.1$\pm$16.7  
\\ 

\bottomrule
    \end{tabular}
    \caption{\textbf{Evaluation results on Guideline 5 - Computational efficiency}. We report the mean $\pm$ std speed in seconds to generate a heatmap on a data point.
    ``NaN'' in the knee task is because the XAI method was not included in the evaluation. The XAI methods are in alphabetic order.
    }
    \label{tab:eval_result_g5}
\end{table}

The computational time spent in generating a heatmap is shown in Table~\ref{tab:eval_result_g5}. The speed  of generating a heatmap was stable across the three datasets with different image dimensions (2D and 3D) and model architectures. Some gradient-based methods that rely solely on backpropagation can generate near real-time explanations, which enables their clinical use in real-time interactive XAI systems. For some gradient-based and all perturbation-based methods that require multiple sampling, their speed is $>10$ seconds or even longer. Methods such as Lime or Shapley Value Sampling need to take 7$\sim$30 minutes to generate a heatmap. Depending on the specific use case and XAI method parameter settings, the long wait time may prevent their clinical use.

\section{Discussion}\label{discussion}
\subsection{Evaluated heatmap methods failed to meet the Clinical XAI Guidelines}

We conducted a systematic evaluation on 16 commonly-used heatmap methods following the Clinical XAI Guidelines. Although the heatmap explanations were easily understandable to clinical users (G1), they only partially fulfilled G2 clinical relevance, due to the missing descriptions of feature pathology from the heatmap, which corresponds to the clinical image interpretation process (\S\ref{g1g2}). The examined heatmap methods did not reliably exhibit the property of G3 Truthfulness on multiple models in the two clinical tasks. Due to the failure of G3, G4 testing for informative plausibility also had poor performance. Most heatmaps were computationally efficient regarding G5 that can generate a heatmap within seconds, except for some sampling-based methods such as Shapley Value Sampling, which may take more than 20 minutes.

Next, we discuss the computational evaluation results on G3 and G4 by referring to the literature, and discuss potential research directions and open research questions. 
\subsubsection{G3 Truthfulness}
In G3, we evaluated whether the examined heatmaps can correctly reveal important features for model decision process at both the coarse-grained modality level and fine-grained feature level. None of the examined XAI methods fulfilled G3 on both glioma and knee tasks. Our findings join a number of previous literature findings on the untruthfulness of post-hoc XAI methods in natural image and MIA tasks~\citep{adebayo2022post, 10.5555/3495724.3495784, NEURIPS2018_294a8ed2, DBLP:journals/corr/abs-2104-14403}, in which they used modified datasets with known ground truth of important features to diagnose spurious or biased features learned by the model. Prior literature hypothesized the reason for the untruthfulness of the post-hoc explanation is that post-hoc methods  summarize statistics that may only reveal partial aspects of a model's internal state, and the actual decision process may be scattered throughout the network~\citep{Chen2020}. Therefore, prior work called for inherently interpretable AI models instead in high-stakes domains~\citep{Rudin2019}. Both post-hoc XAI and inherently interpretable AI models require truthfulness assessment~\citep{jacovi-goldberg-2020-towards}.

\subsubsection{G4 Informative plausibility}\label{dis_g4}

In G4, we tested the MSFI correlation with two indicators for model decision quality: \textbf{1}) model output probability, and \textbf{2}) model prediction correctness. For \textbf{1}) model output probability,  on the glioma task, our assessment showed the plausibility measure can be correlated with model prediction probability, which aligns with prior literature finding on XAI evaluation for chest X-ray task~\citep{Saporta2021.02.28.21252634}. 
For \textbf{2}) testing informative plausibility using model prediction correctness, our results showed existing post-hoc XAI methods can hardly reveal information on model decision correctness, on both the glioma and knee task. This echoes with prior literature finding on a chest X-ray task that showed 
no strong correspondence between model generalization performance and heatmap plausibility measure~\citep{viviano2021saliency}.

The above findings indicate that existing post-hoc heatmap methods may be able to reveal information that is obvious, or \textit{known} to the model (such as the prediction label and its probability), but not good at revealing information that is difficult to estimate, or \textit{unknown} to the model (such as prediction correctness, quality, or reliability). The former information on prediction probability is straightforward for clinical users to obtain by reading the model output, without the extra effort to interpret and assess its explanation; whereas the latter information on decision quality has more clinical significance as shown in our user study (U2. Clinical utility of explainable AI), and is more relevant to the clinical users to spend extra time interpreting the explanation and assessing its plausibility. 

Generating explanations that can be informative for model decision quality is a challenging and clinically important problem. This problem is closely related to uncertainty estimation (UE) for deep learning models~\citep{10.5555/3045390.3045502} that estimates model decision uncertainty.
Compared to providing users with a UE number, generating informative explanations for model decision quality can provide more contextual information to help users understand why, how, and when AI works and does not work.
Despite its clinical importance, proposing and evaluating XAI for model decision verification (G4 Informative plausibility) is an underexplored problem, and there are only a few works~\citep{NEURIPS2021_4e246a38, 10.1007/978-3-030-59710-8_77, Patro_2019_ICCV} that combine UE with XAI by bringing a probabilistic Bayesian view to XAI algorithms. But these proposed XAI methods did not incorporate plausibility measure as a way to quantify explanation uncertainty and its corresponding model decision uncertainty, and their ability to fulfill G4 on revealing model decision quality with plausibility measure is unknown and not assessed. Our Clinical XAI Guidelines and evaluation propose this open and clinically important problem to the research community.

\subsection{Comparison of the guideline criteria}\label{g2vsg4}
Both G1 Understandability and G2 Clinical relevance are qualitative assessments with respect to clinical applicability of the general \textit{form} of an explanation, and are non-specific to an XAI method and the content it generated. In contrast, the other guidelines, G3 Truthfulness, G4 Informative plausibility, and G5 Computational efficiency are quantitative and computational assessments of the explanation \textit{content}, and are specific to each XAI method that generates the explanation content within a specific explanation form. The explanation form can be regarded as different modalities of the explanation information, such as explaining using features, examples, or rules. 
Whereas the explanation content is the specific information expressed through an explanation form. Moreover, although G4 Informative plausibility and G2 Clinical relevance both focus on the aspect of human interpretation of the explanation, plausibility focuses on the content of explanation, whereas G2 Clinical relevance assesses a group of XAI methods that are represented in the same explanation form. An explanation that has a high score in G4 may not be clinically relevant (G2). For example, the content of a heatmap assessed by a plausibility measure may be very indicative of model decision quality,
thus it has a high score for G4.
But the general \textit{form} of heatmap is not completely clinically relevant (G2), because it only provides localization information without information on feature pathology (as detailed in Section~\ref{g1g2}). Similarly, an explanation that is clinically relevant (G2) may not always correspond to a high score in G4. For example, if a group of XAI algorithms provides information on both feature localization and pathology identification, they are considered to be clinically relevant (G2). Within this group, different XAI algorithms may have different performances on their G4 scores, depending on how well the explanation plausibility correlates with AI decision quality.

Since G1 Understandability and G2 Clinical relevance assess the explanation form, an explanation form that passed G1 and G2 can be used to select or propose a group of XAI algorithms that generate the same form. For example, our user study discovered a clinically relevant explanation form of feature attribution: an explanation should at least present feature information on localization and pathology description (\S\ref{g2}). This may cover the explanation form of segmentation maps labeled with different pathology~\citep{DeFauw2018}, or a heatmap coupled with pathological description. Any XAI algorithms that generate such explanation forms are considered to fulfill G2. Some user studies have examined or identified explanation forms on understandability~\citep{jin2021euca, 10.1145/3301275.3302289, 10.1145/3290605.3300234} and clinical relevance~\citep{jin-doctor-user-study}. User studies like these may enable AI developers to bypass G1 or G2 assessment by directly applying the relevant user study findings from the literature to their individual tasks. They can also serve as a starting point for the clinical AI development team before communicating with clinical users to assess G1 and G2.

Much of the literature on XAI evaluation considers the plausibility measure as a requirement~\citep{10.1007/978-3-030-63419-3_3, DESOUZA2021104578, Saporta2021.02.28.21252634, Arun2021}. The Clinical XAI Guidelines do not include the stand-alone plausibility as a clinical requirement, because G1 Understandability and G2 Clinical relevance already regulate an XAI to be clinically viable in its explanation \textit{form}, and the explanation \textit{content} itself does not necessarily need to align with human knowledge (measured by plausibility). Instead of making an explanation plausible to users to gain their trust with a shortcut (i.e., by bypassing the G3 Truthfulness assessment), the Clinical XAI Guidelines focus on the clinical utility of user's plausibility assessment, and inspect whether users' plausibility assessment can shed light on the downstream clinical utilities (U2. Clinical utility of explainable AI), and help users answer their questions following their plausibility assessment (G4 Informative plausibility), such as enabling users to verify model decision, to diagnose model decision flaws and biases, or to discover new knowledge. All these utilities do not require the explanation content to align with human prior knowledge. In fact, we argue that it may be dangerous to select or optimize an XAI method solely on the basis of its plausibility measure. As observed in our user study and in prior literature~\citep{DBLP:journals/corr/abs-2006-04948}, a potential consequence is that the XAI method may be optimized to deceive users and make them overtrust a wrong AI decision with its seemingly plausible explanation, rather than help users to verify the decision quality.

\subsection{Use of the Clinical XAI Guidelines}

Our systematic evaluation demonstrated the use of the guidelines in the evaluation of XAI in two clinical tasks.
Specifically, if we go back to Alex's questions in the beginning, to apply the guidelines to a clinical XAI problem for XAI method selection or proposal, AI designers like Alex may first talk to their target clinical users or other stakeholders to understand their AI literacy (G1 Understandability), their clinical reasoning process which relates to the interpretation of explanation (G2 Clinical relevance). Based on the conversation, AI designers may have a clearer idea about which form(s) of explanation to target. 

For the targeted form of explanation such as feature attribution map, there may be multiple XAI algorithms that can generate it. To design or select the optimal XAI algorithm of the target explanation form, AI designers may choose suitable metrics to assess and optimize XAI methods on the G3 Truthfulness measure. AI designers may also need to test the truthfulness metrics for an XAI algorithm on multiple trained AI models to examine the robustness of XAI method in truly reflecting the model decision process.

For the XAI method candidates that passed the truthfulness assessment, to validate whether the explanation is clinically useful in alerting physicians to AI potential decision flaws, AI developers may further test such property for the XAI method candidates (G4 Informative plausibility). To do so, AI designers can ask clinical users about which features or criteria they are based on to judge the plausibility of explanation, and select computational metrics and prepare data annotations based on the plausibility quantification criteria. Then AI developers can test the correlation between plausibility and decision quality.

AI designers may also need to record the G5 Computational efficiency of the XAI method candidates to rule out the ones that do not meet the speed and computational resource requirement in clinical deployment.

\section{Limitations and future work}
The Clinical XAI Guidelines focus on the general clinical requirements for AI explanation. Some task-dependent requirements for XAI methods, such as data privacy protection, were not included in the guidelines. They can serve as add-on requirements in addition to the guideline criteria for specific clinical tasks.

Our evaluation provides a demonstration of the XAI assessment process to align with clinical requirements. We modified existing methods or proposed ours for the assessment of G3 and G4,
and we do not claim that they are the best evaluation methods for the general guideline criteria. We list the limitations for each evaluation method below:

For G3 Truthfulness: \textbf{1}) Cumulative feature removal experiment has a feature independence assumption, which is violated in image data setting; and there is no consensus on how to set feature replacement value that can keep the same data distribution and not introduce additional information~\citep{frye2021shapley, DBLP:journals/corr/abs-2105-10719}. \textbf{2}) Modality importance correlation experiment only evaluates important features from a modality as a whole, which is too coarse for MIA settings. \textbf{3}) When using synthetic or modified datasets with known ground truth of important features to evaluate XAI methods, it is unknown how well we can generalize the conclusion from the synthetic to real-patient task, given the model and data distribution discrepancies between the two. 

For G4 Informative plausibility: the statistical test for informative plausibility requires the number of wrongly predicted test data to reach a certain sample size for statistical power, which may be difficult to acquire with a highly accurate model and small test set. The statistical test does not identify whether the plausibility measure of correctly and incorrectly predicted data are well separated, and we had to manually visualize the data distribution. 

Future work may propose novel XAI evaluation methods and automated, end-to-end, standardized evaluation pipeline corresponding to the guidelines to speed up the clinical development of XAI techniques.

\section{Conclusion}
In this work, we propose the Clinical XAI Guidelines to support the design and evaluation of clinically-oriented XAI systems. The proposal of the guidelines was based on dual understandings of the clinical requirements for explanations from our physician user study, and technical understanding from our previous XAI evaluation studies and XAI literature. The guidelines G1 Understandability and G2 Clinical relevance provide clinical insights for the selection of explanation forms. Guidelines G3 Truthfulness, G4 Informative plausibility, and G5 Computational efficiency incorporate the clinical requirements for explanation as clear technical objectives to be optimized for.

Based on the guidelines, we conducted a systematic evaluation on 16 commonly-used heatmap methods. The evaluation focused on a technically-novel and clinically-pervasive problem of multi-modal medical image explanation with two clinical tasks of brain tumor grading and knee lesion identification. 
We proposed a novel metric, MSFI for multi-modal medical image explanation tasks, to bypass physicians' manual assessment of explanation plausibility. The evaluation results showed that the evaluated heatmap methods failed to fulfill G3 and G4, thus were not suitable for clinical use. The evaluation demonstrates the use of Clinical XAI Guidelines in real-world clinical tasks to facilitate the design and evaluation of clinically-oriented XAI.

\section*{Acknowledgments}
We thank all physician participants in our user study. We thank Sunho Kim, Yiqi Yan, Mayur Mallya, Shahab Aslani, Ben Cardoen, and Hanene Ben Yedder for their technical support and helpful discussions. We thank the reviewers for their time, efforts, and insightful comments.
This research was supported by the BC Cancer Foundation-BrainCare Fund, and Borealis AI through the Borealis AI Global Fellowship Award. This research was also enabled in part by the computational resources provided by NVIDIA and the Digital Research Alliance of Canada (alliancecan.ca).

\section*{Conflicts of interest}
None.

\section*{Code availability}
Code is available at: \\
\href{http://github.com/weinajin/multimodal_explanation}{github.com/weinajin/multimodal\_explanation}

\section*{Supplementary material}
Supplementary Material S1 and S2 are available at: \\
\href{https://github.com/weinajin/multimodal_explanation/tree/main/paper}{github.com/weinajin/multimodal\_explanation/tree/main/paper}

\section*{Appendix\\\vspace{0.5em}Clinical Explainable AI Guidelines (Full Version)}\label{appendix}

In an effort to guide the design and evaluation of clinical XAI to meet both clinical and technical requirements, we present a checklist including five canonical criteria which we believe may serve as guidelines for developing clinically-oriented XAI. The guidelines were developed with a collective effort from both clinical and technical aspects with complementary expertise in AI, human factor analysis, and clinical practice. In addition, it was driven and supported by the findings from our physician user study, pilot XAI evaluation experiments~\citep{aaai2022,DBLP:journals/corr/abs-2107-05047}, and literature.
We sought feedback from two physicians and several researchers on medical image analysis as a heuristic evaluation of the guidelines.

To acquire physicians' requirements for clinical XAI, we conducted a physician user study with 30 neurosurgeons to elicit their clinical requirements by using a clinical XAI prototype. The low-fidelity prototype is a clinical decision-support AI system that provides suggestions from a CNN model to differentiate lower-grade gliomas from high-grade ones based on multi-modal MRI. For each AI suggestion, it also shows a heatmap explanation that highlights the important features for model prediction. The user study consisted of an online survey that embedded the XAI prototype and collected physicians' quantitative ratings of the heatmaps, and an optional post-survey interview where physicians comment on the clinical XAI system. Five physicians participated in the interview, and seven physicians provided comments in the survey by answering open-ended questions. We analyzed the qualitative data collected from interview sessions and open-ended questions in the survey as the main support to develop the guidelines from the clinical aspect. The detailed user study findings and method are in Supplementary Material S1, and its related supporting sections were referred to in the guidelines starting with `U'.

Next, we present the Clinical XAI Guidelines, which consist of five evaluation objectives to optimize a clinical XAI technique. They are categorized into three considerations on clinical usability, evaluation, and operation. For each objective in the guidelines, we list its key references from our user study or literature. We also analyze examples that follow the objective and/or counterexamples that violate it. Ways of assessment are also described to help identify if the objective is met. The guidelines and their key points are summarized in Table~\ref{table:gl}.

\subsection{Clinical usability considerations}
\noindent \textbf{Guideline 1: Understandability.}

The form and context of an explanation should be easily understandable by its clinical users. Users do not need to have technical knowledge in machine learning, AI, or programming to interpret the explanation.

\begin{itemize}
    \item \textbf{Example}:
    
Physicians find the feature attribution maps (heatmaps) used in our user study easily understandable. Other explanation forms on medical image analysis tasks such as similar examples~\citep{10.1145/3290605.3300234}, counterfactual examples~\citep{10.1007/978-3-030-32226-7_76},
scoring (linear feature attribution)~\citep{8333693}, or rule-based explanation, are shown in prior physician user studies in the literature. \cite{jin2021euca} summarized 12 end-user-friendly explanation forms that do not require technical knowledge, including feature-based (feature attribution, feature shape, feature interaction), example-based (similar, prototypical, and counterfactual example), rule-based explanation (rules, decision tree), and contextual information (input, output, performance, dataset). In addition to the explanation that reveals the model decision process, in our user study, physicians also required other information that makes the AI model transparent, such as model performance, training dataset, and prediction confidence (U3.3. Making AI transparent by providing information on performance, training dataset, and decision confidence). An XAI system may use one or a combination of multiple explanation forms that are friendly to clinical users.

\item \textbf{Counterexample}: 

A counterexample of understandability is to explain by visualizing the learned representation of neurons in DNN~\citep{Olah2017}. Although the \textit{form} of neuron visualization as images is intuitive to look at, interpreting the images requires users to have prior knowledge on DNN model and neuron to understand the \textit{context} of neuron visualization. 

\item \textbf{Assessment method}:

To assess if the understandability objective is met, AI designers can conduct a self-assessment on an XAI technique to inspect its AI knowledge prerequisites, conduct a pilot physician usability study using low-fidelity prototypes, or have informal conversations with clinical users to understand their minimal AI literary, and choose proper explanation techniques accordingly. Low-fidelity prototypes such as sketches can be used as a quick trial-and-error tool and help clinical users better vision an explanation in a clinical context. As a reference, ~\cite{jin2021euca} provides users' understandability from 32 laypersons on 12 end-user-friendly explanation forms, and prototyping support to identify clinical user-friendly explanations. This assessment is usually one-time, conducted at the initial phase of a project.

\end{itemize}
\noindent \textbf{Guideline 2: Clinical relevance.}

The way physicians use explanations is to inspect the AI-based evidence provided by the explanation, and incorporate such evidence in their clinical reasoning process for downstream tasks, such as assessing the validity of AI decision, making a final decision on the case, improving their problem-solving skills, or making scientific discoveries (U2. Clinical utility of explainable AI; U1. Clinical utility of AI).
To make XAI clinically useful, the explanation information should be relevant to physicians' clinical decision-making pattern, and can support their clinical reasoning process.

For diagnostic/predictive tasks on clinical images, physicians' image interpretation process includes two general steps: \textbf{1}) feature extraction: physicians first perform pattern recognition to localize key features and identify pathology of these features; \textbf{2}) reasoning on the extracted features: physicians perform medical reasoning and construct diagnostic hypotheses (differential diagnosis) based on the image feature evidence. A clinically relevant explanation 
should provide information corresponding to the above
process, so that physicians can incorporate the explanation information into their medical image interpretation process (U3. Clinical requirements of explainable AI). 

\begin{displayquote}
\textit{``What (explanation) we get currently, when a radiologist read it, they point out the significant features, and then they integrate those knowledge, and say, to my best guess, this is a GBM. And I have the same expectations of AI (explanation).''} (N3)
\end{displayquote}

\begin{itemize}
    \item  \textbf{Example}: 

In the user study, physicians visioned the ideal explanations that are clinically relevant (U3.2. Desirable explanation), such as using radiologists' language, a linear scoring model, or a rule-based explanation. Those explanations are composed of clinically meaningful features. And their form of text, rule, or linear model corresponds to the second step of the reasoning process on the extracted features in the above clinical image interpretation process.

\item \textbf{Counterexample}: 

The heatmap explanation is not completely clinically relevant, as physicians were perplexed by the highlighted areas, regardless of whether the areas align with their prior knowledge or not. Because the heatmap explanation only performs half of the clinical image interpretation step 1) of feature localization, it lacks the description of important features, let alone to perform reasoning on these features (U3.1. Limitations of existing heatmap explanation).

\begin{displayquote}
\textit{``Though the heatmap is drawing your eyes to many different spots, but I feel like I didn't understand why my eyes were being driven to those spots, like why were these very specific components important? And I think that's where all my confusion was.''} (N2)
\end{displayquote}

\item \textbf{Assessment method}:

A user study with the target clinical users can be conducted in a formal or informal manner, to understand the clinical decision-making pattern or workflow for the target task, and inspect whether the explanation form corresponds to such pattern, and can help physicians answer their questions on the rationale of the model decision, how do users incorporate the explanation information into their decision process. The above information can be collected via an interview or conversation with users, a field visit and observation, or a focus group, etc. Low-fidelity prototypes (such as sketches)~\citep{jin2021euca} of explanation form candidates can be used to elicit more in-context feedback from clinical users' communication. The G2 assessment can be co-conducted with G1 assessment at the initial phase of a project, and it is also a one-time assessment. As a reference, our user study finding (U2 and U3 in Supplementary Material S1) provides G2 assessment results for the explanation form of heatmap.
\end{itemize}

\subsection{Evaluation considerations}

\noindent \textbf{Guideline 3: Truthfulness.}

Explanation should truthfully reflect the model decision process. This is the fundamental requirement for a clinically-oriented explanation, and an explanation method should fulfill the truthfulness requirement first prior to other evaluation requirements such as G4: Informative plausibility in the guidelines.

\begin{itemize}
    \item \textbf{Counterexample}: 

One of the main clinical utilities of explanation is that clinical users intuitively use explanation plausibility assessment (G4) to verify AI decisions for a case to decide whether to take or reject the AI suggestion, and calibrate their trust in AI's current prediction on the case, or the AI model in general accordingly (U2.3). Users do so with an implicit assumption that explanations are the true representation of the model decision process. 
Violating truthfulness can lead to two significant consequences during the human assessment on explanation plausibility (G4): 

\textbf{1}. Clinical users may mistakenly reject AI's correct suggestion merely for the poor performance of the XAI method, which shows an unreasonable explanation.

\textbf{2}. If an XAI method is proposed or selected based on explanation plausibility objective only, rather than help clinical users to verify the decision quality, the explanation can be optimized to deceive clinical users with its seemingly plausible explanation, despite the wrong prediction from AI~\citep{DBLP:journals/corr/abs-2006-04948},
as illustrated by the physician participant N1's quote:

\begin{displayquote}
\textit{``If a system made its prediction based upon these areas (outside the tumor), I would definitely not trust that system, but I would be very reassured that the system is telling me that. ...So I’m less likely to use this model, but I’m more likely to use a model that does a better job than this, because I am reassured that when I see that better model, that I will be able to have access to that back-end explanation. ''} (N1)
\end{displayquote}

\item \textbf{Assessment method}:

As stated in~\citep{jacovi-goldberg-2020-towards}, the truthfulness or faithfulness objective cannot and should not be assessed by human judgment on the explanation quality or annotations of the human prior knowledge, because humans do not know the model's underlying decision process. 

The most common way to assess explanation truthfulness for feature attribution XAI methods in the literature is to gradually add or remove features from the most to the least important ones according to an explanation, and measure the model performance change~\citep{DBLP:journals/corr/abs-2104-08782,NEURIPS2019_a7471fdc,DBLP:conf/nips/HookerEKK19,7552539,Lundberg2020,10.5555/3327757.3327875}. Another way is to construct synthetic evaluation datasets in which the ground truth knowledge on the model decision process from input features to prediction is known and controlled~\citep{doshivelez2017rigorous,pmlr-v80-kim18d,DBLP:journals/corr/abs-1806-00069}.
\end{itemize}

\noindent \textbf{Guideline 4: Informative plausibility.}

The ultimate use of an explanation is to be interpreted and assessed by clinical users. Physicians intuitively use the assessment of explanation plausibility or reasonableness (i.e.: how reasonable the explanation is based on its agreement with human prior knowledge on the task) as a way to evaluate AI decision quality, so that to achieve multifaceted clinical utilities with XAI, including verifying AI's decisions (U2.3), calibrating trust in AI (U2.3), ensuring the safe use of AI, resolving disagreement with AI (U2.2), identifying potential biases, and making medical discoveries (U2.4). Informative plausibility aims to validate whether an XAI method can achieve its utility in helping users to identify potential AI decision flaws and/or biases, i.e.: a plausible explanation for a right decision, and an implausible explanation for a wrong decision of AI. G3 Truthfulness is the gatekeeper of G4 Informative plausibility
to warrant the explanation truthfully represents the AI decision process.

\begin{itemize}
    \item \textbf{Example}: 

In our evaluation, we abstract physicians' clinical requirements for multi-modal medical image explanation (U4) into the MSFI metric. It regards the most plausible heatmap explanation as some maps that can both localize the important image feature on each imaging modality, and highlight the important modalities for decision. We evaluate how well MSFI metric corresponds to physicians' assessment by quantitative measure to calculate the correlation between the two, and showcase the visual examples as a qualitative measure. We then inspect the subsequent utility of the MSFI metric on verifying model decisions, by measuring its correlation with decision correctness.

\item \textbf{Assessment method}:

To test whether explanation plausibility is informative to help users identify AI decision errors and biases, AI designers can assess the correlation between AI decision quality measures (such as model performance, calibrated prediction uncertainty, prediction correctness, and quantification of biased patterns) with plausibility measures~\citep{adebayo2022post, Saporta2021.02.28.21252634}. 

Since human assessment of explanation plausibility is usually subjective and susceptible to biases (U5.2. Bias and limitation of physicians’ quantitative rating), AI designers may consider quantifying the plausibility measure by abstracting the human assessment criteria into computation metrics for a given task. The quantification of human assessment is \textit{not} meant to directly select or optimize XAI methods for clinical use. Rather, XAI methods should be optimized for their truthfulness measures (G3). Plausibility quantification is meant to validate the capability of XAI methods on their subsequent clinical utility to reveal AI decision flaws and/or biases,
providing their high truthfulness score. Quantifying plausibility can make such a validation process automatic, reproducible, standardizable, and computationally efficient. 
Similarly, the human annotation of important features according to physicians' prior knowledge, which is used to quantify plausibility, cannot be regarded as the ``ground truth'' of explanation, because explanations (given that they fulfill G3 Truthfulness) are still acceptable even if they are not aligned with human prior knowledge, but reveal the model decision quality or help humans to identify new patterns and make medical discoveries.

Many approaches were proposed to quantify explanation plausibility measure. These measures calculate the agreement of explanation with human prior knowledge annotations for a given task~\citep{10.1007/978-3-030-32226-7_82,netdissect2017,Arun2021}. 
To evaluate whether the quantified plausibility measure is a good substitute for human assessment, AI designers can use a quantitative measure by calculating the correlation between the plausibility metric and clinical users' assessment score, or use a qualitative measure by showing physicians different explanations and their plausibility score, and ask them to judge.

\end{itemize}
\subsection{Operational consideration}
\noindent \textbf{Guideline 5: Computational efficiency.}

Since many AI-assisted clinical tasks are time-sensitive decisions (U1.2.1. Decision support for time-sensitive cases, and hard cases), the selection or proposal of clinical XAI techniques needs to consider the computational time and resources. The wait time for an explanation should not be a bottleneck for the clinical task workflow. 

\begin{itemize}
    \item  \textbf{Example}: 

In our evaluation, some gradient-based XAI methods that use backpropagation can generate near real-time explanations with an upper limit of up to 10 seconds. This also enables their clinical use in generating real-time interactive explanations.

\item \textbf{Counterexample}: 

For XAI techniques that require sampling input-output pairs, 
their computational time may be too long for physicians to wait for an explanation. In our evaluation, it took about 30 minutes for Shapley Value Sampling method to generate one heatmap on a typical desktop computer with GPU. 

\item \textbf{Assessment method}:

AI designers can record the computational time and resources for XAI method to assess whether the requirement of computational efficiency is met.
AI designers may also need to talk to clinical users to understand whether their clinical task includes time-sensitive decisions, and their maximum tolerable waiting time for an explanation on the task. For some XAI methods, the computational time depends on the settings of some specific parameters, such as the number and size of feature masks to generate the perturbed samples, and the number of samples. AI designers need to identify the optimal set of parameters to balance explanation accuracy and computational efficiency.
\end{itemize}

\clearpage
\bibliography{xai_eval,xai_eval_202111}

\begin{thebibliography}{117}
\providecommand{\natexlab}[1]{#1}
\providecommand{\url}[1]{\texttt{#1}}
\expandafter\ifx\csname urlstyle\endcsname\relax
  \providecommand{\doi}[1]{doi: #1}\else
  \providecommand{\doi}{doi: \begingroup \urlstyle{rm}\Url}\fi

\bibitem[Adebayo et~al.(2018)Adebayo, Gilmer, Muelly, Goodfellow, Hardt, and
  Kim]{NEURIPS2018_294a8ed2}
J.~Adebayo, J.~Gilmer, M.~Muelly, I.~Goodfellow, M.~Hardt, and B.~Kim.
\newblock Sanity checks for saliency maps.
\newblock In S.~Bengio, H.~Wallach, H.~Larochelle, K.~Grauman, N.~Cesa-Bianchi,
  and R.~Garnett, editors, \emph{Advances in Neural Information Processing
  Systems}, volume~31. Curran Associates, Inc., 2018.
\newblock URL
  \url{https://proceedings.neurips.cc/paper/2018/file/294a8ed24b1ad22ec2e7efea049b8737-Paper.pdf}.

\bibitem[Adebayo et~al.(2020)Adebayo, Muelly, Liccardi, and
  Kim]{10.5555/3495724.3495784}
J.~Adebayo, M.~Muelly, I.~Liccardi, and B.~Kim.
\newblock Debugging tests for model explanations.
\newblock In \emph{Proceedings of the 34th International Conference on Neural
  Information Processing Systems}, NIPS'20, Red Hook, NY, USA, 2020. Curran
  Associates Inc.
\newblock ISBN 9781713829546.

\bibitem[Adebayo et~al.(2022)Adebayo, Muelly, Abelson, and
  Kim]{adebayo2022post}
J.~Adebayo, M.~Muelly, H.~Abelson, and B.~Kim.
\newblock Post hoc explanations may be ineffective for detecting unknown
  spurious correlation.
\newblock In \emph{International Conference on Learning Representations}, 2022.
\newblock URL \url{https://openreview.net/forum?id=xNOVfCCvDpM}.

\bibitem[Alvarez-Melis and Jaakkola(2018)]{10.5555/3327757.3327875}
D.~Alvarez-Melis and T.~S. Jaakkola.
\newblock Towards robust interpretability with self-explaining neural networks.
\newblock In \emph{Proceedings of the 32nd International Conference on Neural
  Information Processing Systems}, NIPS'18, page 7786–7795, Red Hook, NY,
  USA, 2018. Curran Associates Inc.

\bibitem[Alvarez{-}Melis and Jaakkola(2018)]{DBLP:journals/corr/abs-1806-08049}
D.~Alvarez{-}Melis and T.~S. Jaakkola.
\newblock On the robustness of interpretability methods.
\newblock \emph{CoRR}, abs/1806.08049, 2018.
\newblock URL \url{http://arxiv.org/abs/1806.08049}.

\bibitem[Amann et~al.(2020)Amann, Blasimme, Vayena, Frey, and Madai]{Amann2020}
J.~Amann, A.~Blasimme, E.~Vayena, D.~Frey, and V.~I. Madai.
\newblock {Explainability for artificial intelligence in healthcare: a
  multidisciplinary perspective}.
\newblock \emph{BMC Medical Informatics and Decision Making}, 20\penalty0
  (1):\penalty0 310, dec 2020.
\newblock ISSN 14726947.
\newblock \doi{10.1186/s12911-020-01332-6}.
\newblock URL
  \url{https://bmcmedinformdecismak.biomedcentral.com/articles/10.1186/s12911-020-01332-6}.

\bibitem[Arun et~al.(2021)Arun, Gaw, Singh, Chang, Aggarwal, Chen, Hoebel,
  Gupta, Patel, Gidwani, Adebayo, Li, and Kalpathy-Cramer]{Arun2021}
N.~Arun, N.~Gaw, P.~Singh, K.~Chang, M.~Aggarwal, B.~Chen, K.~Hoebel, S.~Gupta,
  J.~Patel, M.~Gidwani, J.~Adebayo, M.~D. Li, and J.~Kalpathy-Cramer.
\newblock Assessing the trustworthiness of saliency maps for localizing
  abnormalities in medical imaging.
\newblock \emph{Radiology: Artificial Intelligence}, 3\penalty0 (6), Nov. 2021.
\newblock \doi{10.1148/ryai.2021200267}.
\newblock URL \url{https://doi.org/10.1148/ryai.2021200267}.

\bibitem[Bakas et~al.(2017)Bakas, Akbari, Sotiras, Bilello, Rozycki, Kirby,
  Freymann, Farahani, and Davatzikos]{Bakas2017}
S.~Bakas, H.~Akbari, A.~Sotiras, M.~Bilello, M.~Rozycki, J.~S. Kirby, J.~B.
  Freymann, K.~Farahani, and C.~Davatzikos.
\newblock Advancing the cancer genome atlas glioma {MRI} collections with
  expert segmentation labels and radiomic features.
\newblock \emph{Scientific Data}, 4\penalty0 (1), Sept. 2017.
\newblock \doi{10.1038/sdata.2017.117}.
\newblock URL \url{https://doi.org/10.1038/sdata.2017.117}.

\bibitem[{Barredo Arrieta} et~al.(2020){Barredo Arrieta}, Díaz-Rodríguez,
  {Del Ser}, Bennetot, Tabik, Barbado, Garcia, Gil-Lopez, Molina, Benjamins,
  Chatila, and Herrera]{BARREDOARRIETA202082}
A.~{Barredo Arrieta}, N.~Díaz-Rodríguez, J.~{Del Ser}, A.~Bennetot, S.~Tabik,
  A.~Barbado, S.~Garcia, S.~Gil-Lopez, D.~Molina, R.~Benjamins, R.~Chatila, and
  F.~Herrera.
\newblock Explainable artificial intelligence (xai): Concepts, taxonomies,
  opportunities and challenges toward responsible ai.
\newblock \emph{Information Fusion}, 58:\penalty0 82--115, 2020.
\newblock ISSN 1566-2535.
\newblock \doi{https://doi.org/10.1016/j.inffus.2019.12.012}.
\newblock URL
  \url{https://www.sciencedirect.com/science/article/pii/S1566253519308103}.

\bibitem[Bau et~al.(2017)Bau, Zhou, Khosla, Oliva, and
  Torralba]{netdissect2017}
D.~Bau, B.~Zhou, A.~Khosla, A.~Oliva, and A.~Torralba.
\newblock Network dissection: Quantifying interpretability of deep visual
  representations.
\newblock In \emph{Computer Vision and Pattern Recognition}, 2017.

\bibitem[Bello et~al.(2019)Bello, Dawes, Duan, Biffi, de~Marvao, Howard, Gibbs,
  Wilkins, Cook, Rueckert, and O'Regan]{Bello2019}
G.~A. Bello, T.~J.~W. Dawes, J.~Duan, C.~Biffi, A.~de~Marvao, L.~S. G.~E.
  Howard, J.~S.~R. Gibbs, M.~R. Wilkins, S.~A. Cook, D.~Rueckert, and D.~P.
  O'Regan.
\newblock {Deep-learning cardiac motion analysis for human survival
  prediction}.
\newblock \emph{Nature Machine Intelligence}, 1\penalty0 (2):\penalty0 95--104,
  2019.
\newblock ISSN 2522-5839.
\newblock \doi{10.1038/s42256-019-0019-2}.
\newblock URL \url{https://doi.org/10.1038/s42256-019-0019-2}.

\bibitem[Beyer et~al.(2002)Beyer, Townsend, and Blodgett]{pmid12072843}
T.~Beyer, D.~W. Townsend, and T.~M. Blodgett.
\newblock {{D}ual-modality {P}{E}{T}/{C}{T} tomography for clinical oncology}.
\newblock \emph{Q J Nucl Med}, 46\penalty0 (1):\penalty0 24--34, Mar 2002.

\bibitem[Bien et~al.(2018)Bien, Rajpurkar, Ball, Irvin, Park, Jones, Bereket,
  Patel, Yeom, Shpanskaya, Halabi, Zucker, Fanton, Amanatullah, Beaulieu,
  Riley, Stewart, Blankenberg, Larson, Jones, Langlotz, Ng, and
  Lungren]{Bien2018}
N.~Bien, P.~Rajpurkar, R.~L. Ball, J.~Irvin, A.~Park, E.~Jones, M.~Bereket,
  B.~N. Patel, K.~W. Yeom, K.~Shpanskaya, S.~Halabi, E.~Zucker, G.~Fanton,
  D.~F. Amanatullah, C.~F. Beaulieu, G.~M. Riley, R.~J. Stewart, F.~G.
  Blankenberg, D.~B. Larson, R.~H. Jones, C.~P. Langlotz, A.~Y. Ng, and M.~P.
  Lungren.
\newblock {Deep-learning-assisted diagnosis for knee magnetic resonance
  imaging: Development and retrospective validation of MRNet}.
\newblock \emph{PLOS Medicine}, 15\penalty0 (11):\penalty0 e1002699, nov 2018.
\newblock ISSN 1549-1676.
\newblock \doi{10.1371/journal.pmed.1002699}.
\newblock URL \url{http://dx.plos.org/10.1371/journal.pmed.1002699}.

\bibitem[Bigolin~Lanfredi et~al.(2019)Bigolin~Lanfredi, Schroeder, Vachet, and
  Tasdizen]{10.1007/978-3-030-32226-7_76}
R.~Bigolin~Lanfredi, J.~D. Schroeder, C.~Vachet, and T.~Tasdizen.
\newblock Adversarial regression training for visualizing the progression of
  chronic obstructive pulmonary disease with chest x-rays.
\newblock In D.~Shen, T.~Liu, T.~M. Peters, L.~H. Staib, C.~Essert, S.~Zhou,
  P.-T. Yap, and A.~Khan, editors, \emph{Medical Image Computing and Computer
  Assisted Intervention -- MICCAI 2019}, pages 685--693, Cham, 2019. Springer
  International Publishing.
\newblock ISBN 978-3-030-32226-7.

\bibitem[Bitar et~al.(2006)Bitar, Leung, Perng, Tadros, Moody, Sarrazin,
  McGregor, Christakis, Symons, Nelson, and Roberts]{Bitar2006}
R.~Bitar, G.~Leung, R.~Perng, S.~Tadros, A.~R. Moody, J.~Sarrazin, C.~McGregor,
  M.~Christakis, S.~Symons, A.~Nelson, and T.~P. Roberts.
\newblock {MR} pulse sequences: What every radiologist wants to know but is
  afraid to ask.
\newblock \emph{{RadioGraphics}}, 26\penalty0 (2):\penalty0 513--537, Mar.
  2006.
\newblock \doi{10.1148/rg.262055063}.
\newblock URL \url{https://doi.org/10.1148/rg.262055063}.

\bibitem[{Bussone} et~al.(2015){Bussone}, {Stumpf}, and {O'Sullivan}]{7349687}
A.~{Bussone}, S.~{Stumpf}, and D.~{O'Sullivan}.
\newblock The role of explanations on trust and reliance in clinical decision
  support systems.
\newblock In \emph{2015 International Conference on Healthcare Informatics},
  pages 160--169, 2015.
\newblock \doi{10.1109/ICHI.2015.26}.

\bibitem[Cai et~al.(2019{\natexlab{a}})Cai, Jongejan, and
  Holbrook]{10.1145/3301275.3302289}
C.~J. Cai, J.~Jongejan, and J.~Holbrook.
\newblock The effects of example-based explanations in a machine learning
  interface.
\newblock In \emph{Proceedings of the 24th International Conference on
  Intelligent User Interfaces}, IUI '19, page 258–262, New York, NY, USA,
  2019{\natexlab{a}}. Association for Computing Machinery.
\newblock ISBN 9781450362726.
\newblock \doi{10.1145/3301275.3302289}.
\newblock URL \url{https://doi.org/10.1145/3301275.3302289}.

\bibitem[Cai et~al.(2019{\natexlab{b}})Cai, Reif, Hegde, Hipp, Kim, Smilkov,
  Wattenberg, Viegas, Corrado, Stumpe, and Terry]{10.1145/3290605.3300234}
C.~J. Cai, E.~Reif, N.~Hegde, J.~Hipp, B.~Kim, D.~Smilkov, M.~Wattenberg,
  F.~Viegas, G.~S. Corrado, M.~C. Stumpe, and M.~Terry.
\newblock Human-centered tools for coping with imperfect algorithms during
  medical decision-making.
\newblock In \emph{Proceedings of the 2019 CHI Conference on Human Factors in
  Computing Systems}, CHI '19, page 1–14, New York, NY, USA,
  2019{\natexlab{b}}. Association for Computing Machinery.
\newblock ISBN 9781450359702.
\newblock \doi{10.1145/3290605.3300234}.
\newblock URL \url{https://doi.org/10.1145/3290605.3300234}.

\bibitem[Cai et~al.(2019{\natexlab{c}})Cai, Winter, Steiner, Wilcox, and
  Terry]{10.1145/3359206}
C.~J. Cai, S.~Winter, D.~Steiner, L.~Wilcox, and M.~Terry.
\newblock "hello ai": Uncovering the onboarding needs of medical practitioners
  for human-ai collaborative decision-making.
\newblock \emph{Proc. ACM Hum.-Comput. Interact.}, 3\penalty0 (CSCW), Nov.
  2019{\natexlab{c}}.
\newblock \doi{10.1145/3359206}.
\newblock URL \url{https://doi.org/10.1145/3359206}.

\bibitem[Carter and Nielsen(2017)]{Carter2017}
S.~Carter and M.~Nielsen.
\newblock Using artificial intelligence to augment human intelligence.
\newblock \emph{Distill}, 2\penalty0 (12), Dec. 2017.
\newblock \doi{10.23915/distill.00009}.
\newblock URL \url{https://doi.org/10.23915/distill.00009}.

\bibitem[Caruana et~al.(2015)Caruana, Lou, Gehrke, Koch, Sturm, and
  Elhadad]{Caruana2015}
R.~Caruana, Y.~Lou, J.~Gehrke, P.~Koch, M.~Sturm, and N.~Elhadad.
\newblock {Intelligible models for healthcare: Predicting pneumonia risk and
  hospital 30-day readmission}.
\newblock In \emph{Proceedings of the ACM SIGKDD International Conference on
  Knowledge Discovery and Data Mining}, volume 2015-Augus, pages 1721--1730,
  New York, New York, USA, aug 2015. Association for Computing Machinery.
\newblock ISBN 9781450336642.
\newblock \doi{10.1145/2783258.2788613}.
\newblock URL \url{http://dl.acm.org/citation.cfm?doid=2783258.2788613}.

\bibitem[Castro et~al.(2009)Castro, Gómez, and Tejada]{CASTRO20091726}
J.~Castro, D.~Gómez, and J.~Tejada.
\newblock Polynomial calculation of the shapley value based on sampling.
\newblock \emph{Computers \& Operations Research}, 36\penalty0 (5):\penalty0
  1726--1730, 2009.
\newblock ISSN 0305-0548.
\newblock \doi{https://doi.org/10.1016/j.cor.2008.04.004}.
\newblock URL
  \url{https://www.sciencedirect.com/science/article/pii/S0305054808000804}.
\newblock Selected papers presented at the Tenth International Symposium on
  Locational Decisions (ISOLDE X).

\bibitem[Chen et~al.(2019)Chen, Li, Tao, Barnett, Su, and
  Rudin]{10.5555/3454287.3455088}
C.~Chen, O.~Li, C.~Tao, A.~J. Barnett, J.~Su, and C.~Rudin.
\newblock \emph{<i>This</i> Looks like <i>That</i>: Deep Learning for
  Interpretable Image Recognition}.
\newblock Curran Associates Inc., Red Hook, NY, USA, 2019.

\bibitem[Chen et~al.(2021)Chen, Chen, Lipkov{\'{a}}, Wang, Williamson, Lu,
  Sahai, and Mahmood]{DBLP:journals/corr/abs-2110-00603}
R.~J. Chen, T.~Y. Chen, J.~Lipkov{\'{a}}, J.~J. Wang, D.~F.~K. Williamson,
  M.~Y. Lu, S.~Sahai, and F.~Mahmood.
\newblock Algorithm fairness in {AI} for medicine and healthcare.
\newblock \emph{CoRR}, abs/2110.00603, 2021.
\newblock URL \url{https://arxiv.org/abs/2110.00603}.

\bibitem[Chen et~al.(2020)Chen, Bei, and Rudin]{Chen2020}
Z.~Chen, Y.~Bei, and C.~Rudin.
\newblock Concept whitening for interpretable image recognition.
\newblock \emph{Nature Machine Intelligence}, 2\penalty0 (12):\penalty0
  772--782, Dec. 2020.
\newblock \doi{10.1038/s42256-020-00265-z}.
\newblock URL \url{https://doi.org/10.1038/s42256-020-00265-z}.

\bibitem[Cochard and Netter(2012)]{cochard_netter_2012}
L.~R. Cochard and F.~H. Netter.
\newblock \emph{Netters introduction to imaging}.
\newblock Elsevier Saunders, 2012.

\bibitem[Critch and Krueger(2020)]{DBLP:journals/corr/abs-2006-04948}
A.~Critch and D.~Krueger.
\newblock {AI} research considerations for human existential safety {(ARCHES)}.
\newblock \emph{CoRR}, abs/2006.04948, 2020.
\newblock URL \url{https://arxiv.org/abs/2006.04948}.

\bibitem[{De Fauw} et~al.(2018){De Fauw}, Ledsam, Romera-Paredes, Nikolov,
  Tomasev, Blackwell, Askham, Glorot, O'Donoghue, Visentin, van~den Driessche,
  Lakshminarayanan, Meyer, Mackinder, Bouton, Ayoub, Chopra, King,
  Karthikesalingam, Hughes, Raine, Hughes, Sim, Egan, Tufail, Montgomery,
  Hassabis, Rees, Back, Khaw, Suleyman, Cornebise, Keane, and
  Ronneberger]{DeFauw2018}
J.~{De Fauw}, J.~R. Ledsam, B.~Romera-Paredes, S.~Nikolov, N.~Tomasev,
  S.~Blackwell, H.~Askham, X.~Glorot, B.~O'Donoghue, D.~Visentin, G.~van~den
  Driessche, B.~Lakshminarayanan, C.~Meyer, F.~Mackinder, S.~Bouton, K.~Ayoub,
  R.~Chopra, D.~King, A.~Karthikesalingam, C.~O. Hughes, R.~Raine, J.~Hughes,
  D.~A. Sim, C.~Egan, A.~Tufail, H.~Montgomery, D.~Hassabis, G.~Rees, T.~Back,
  P.~T. Khaw, M.~Suleyman, J.~Cornebise, P.~A. Keane, and O.~Ronneberger.
\newblock {Clinically applicable deep learning for diagnosis and referral in
  retinal disease}.
\newblock \emph{Nature Medicine}, 24\penalty0 (9):\penalty0 1342--1350, Sept
  2018.
\newblock ISSN 1078-8956.
\newblock \doi{10.1038/s41591-018-0107-6}.
\newblock URL \url{http://www.ncbi.nlm.nih.gov/pubmed/30104768
  http://www.nature.com/articles/s41591-018-0107-6}.

\bibitem[{de Souza} et~al.(2021){de Souza}, Mendel, Strasser, Ebigbo, Probst,
  Messmann, Papa, and Palm]{DESOUZA2021104578}
L.~A. {de Souza}, R.~Mendel, S.~Strasser, A.~Ebigbo, A.~Probst, H.~Messmann,
  J.~P. Papa, and C.~Palm.
\newblock Convolutional neural networks for the evaluation of cancer in
  barrett's esophagus: Explainable ai to lighten up the black-box.
\newblock \emph{Computers in Biology and Medicine}, 135:\penalty0 104578, 2021.
\newblock ISSN 0010-4825.
\newblock
  \doi{[https://doi.org/10.1016/j.compbiomed.2021.104578](https://doi.org/10.1016/j.compbiomed.2021.104578)}.
\newblock URL
  \url{[https://www.sciencedirect.com/science/article/pii/S0010482521003723](https://www.sciencedirect.com/science/article/pii/S0010482521003723)}.

\bibitem[DeYoung et~al.(2020)DeYoung, Jain, Rajani, Lehman, Xiong, Socher, and
  Wallace]{deyoung-etal-2020-eraser}
J.~DeYoung, S.~Jain, N.~F. Rajani, E.~Lehman, C.~Xiong, R.~Socher, and B.~C.
  Wallace.
\newblock {ERASER}: {A} benchmark to evaluate rationalized {NLP} models.
\newblock In \emph{Proceedings of the 58th Annual Meeting of the Association
  for Computational Linguistics}, pages 4443--4458, Online, July 2020.
  Association for Computational Linguistics.
\newblock \doi{10.18653/v1/2020.acl-main.408}.
\newblock URL \url{https://aclanthology.org/2020.acl-main.408}.

\bibitem[Doshi-Velez and Kim(2017)]{doshivelez2017rigorous}
F.~Doshi-Velez and B.~Kim.
\newblock Towards a rigorous science of interpretable machine learning, 2017.

\bibitem[Doshi-Velez and Kim(2018)]{Doshi-Velez2018}
F.~Doshi-Velez and B.~Kim.
\newblock \emph{Considerations for Evaluation and Generalization in
  Interpretable Machine Learning}, pages 3--17.
\newblock Springer International Publishing, Cham, 2018.
\newblock ISBN 978-3-319-98131-4.
\newblock \doi{10.1007/978-3-319-98131-4_1}.
\newblock URL \url{https://doi.org/10.1007/978-3-319-98131-4_1}.

\bibitem[Došilović et~al.(2018)Došilović, Brčić, and Hlupić]{8400040}
F.~K. Došilović, M.~Brčić, and N.~Hlupić.
\newblock Explainable artificial intelligence: A survey.
\newblock In \emph{2018 41st International Convention on Information and
  Communication Technology, Electronics and Microelectronics (MIPRO)}, pages
  0210--0215, 2018.
\newblock \doi{10.23919/MIPRO.2018.8400040}.

\bibitem[Fisher et~al.(2019)Fisher, Rudin, and Dominici]{JMLR:v20:18-760}
A.~Fisher, C.~Rudin, and F.~Dominici.
\newblock All models are wrong, but many are useful: Learning a variable's
  importance by studying an entire class of prediction models simultaneously.
\newblock \emph{Journal of Machine Learning Research}, 20\penalty0
  (177):\penalty0 1--81, 2019.
\newblock URL \url{http://jmlr.org/papers/v20/18-760.html}.

\bibitem[Frye et~al.(2021)Frye, de~Mijolla, Begley, Cowton, Stanley, and
  Feige]{frye2021shapley}
C.~Frye, D.~de~Mijolla, T.~Begley, L.~Cowton, M.~Stanley, and I.~Feige.
\newblock Shapley explainability on the data manifold.
\newblock In \emph{International Conference on Learning Representations}, 2021.
\newblock URL \url{https://openreview.net/forum?id=OPyWRrcjVQw}.

\bibitem[Fujisawa et~al.(2018)Fujisawa, Otomo, Ogata, Nakamura, Fujita,
  Ishitsuka, Watanabe, Okiyama, Ohara, and Fujimoto]{Fujisawa2018}
Y.~Fujisawa, Y.~Otomo, Y.~Ogata, Y.~Nakamura, R.~Fujita, Y.~Ishitsuka,
  R.~Watanabe, N.~Okiyama, K.~Ohara, and M.~Fujimoto.
\newblock Deep-learning-based, computer-aided classifier developed with a small
  dataset of clinical images surpasses board-certified dermatologists in skin
  tumour diagnosis.
\newblock \emph{British Journal of Dermatology}, 180\penalty0 (2):\penalty0
  373--381, Sept. 2018.
\newblock \doi{10.1111/bjd.16924}.
\newblock URL \url{https://doi.org/10.1111/bjd.16924}.

\bibitem[Futoma et~al.(2020)Futoma, Simons, Panch, Doshi-Velez, and
  Celi]{Futoma2020}
J.~Futoma, M.~Simons, T.~Panch, F.~Doshi-Velez, and L.~A. Celi.
\newblock {The myth of generalisability in clinical research and machine
  learning in health care}.
\newblock \emph{The Lancet Digital Health}, 2\penalty0 (9):\penalty0
  e489--e492, sep 2020.
\newblock ISSN 2589-7500.
\newblock \doi{10.1016/S2589-7500(20)30186-2}.
\newblock URL \url{https://doi.org/10.1016/S2589-7500(20)30186-2}.

\bibitem[Gal and Ghahramani(2016)]{10.5555/3045390.3045502}
Y.~Gal and Z.~Ghahramani.
\newblock Dropout as a bayesian approximation: Representing model uncertainty
  in deep learning.
\newblock In \emph{Proceedings of the 33rd International Conference on
  International Conference on Machine Learning - Volume 48}, ICML'16, page
  1050–1059. JMLR.org, 2016.

\bibitem[Gilpin et~al.(2018)Gilpin, Bau, Yuan, Bajwa, Specter, and
  Kagal]{DBLP:journals/corr/abs-1806-00069}
L.~H. Gilpin, D.~Bau, B.~Z. Yuan, A.~Bajwa, M.~A. Specter, and L.~Kagal.
\newblock Explaining explanations: An approach to evaluating interpretability
  of machine learning.
\newblock \emph{CoRR}, abs/1806.00069, 2018.
\newblock URL \url{http://arxiv.org/abs/1806.00069}.

\bibitem[Guidotti et~al.(2018)Guidotti, Monreale, Ruggieri, Turini, Giannotti,
  and Pedreschi]{10.1145/3236009}
R.~Guidotti, A.~Monreale, S.~Ruggieri, F.~Turini, F.~Giannotti, and
  D.~Pedreschi.
\newblock A survey of methods for explaining black box models.
\newblock \emph{ACM Comput. Surv.}, 51\penalty0 (5), aug 2018.
\newblock ISSN 0360-0300.
\newblock \doi{10.1145/3236009}.
\newblock URL \url{https://doi.org/10.1145/3236009}.

\bibitem[Guo et~al.(2017)Guo, Pleiss, Sun, and
  Weinberger]{DBLP:journals/corr/GuoPSW17}
C.~Guo, G.~Pleiss, Y.~Sun, and K.~Q. Weinberger.
\newblock On calibration of modern neural networks.
\newblock \emph{CoRR}, abs/1706.04599, 2017.
\newblock URL \url{http://arxiv.org/abs/1706.04599}.

\bibitem[Harris et~al.(1993)Harris, Adams, Lloyd, and Harvey]{HARRIS1993241}
K.~Harris, H.~Adams, D.~Lloyd, and D.~Harvey.
\newblock The effect on apparent size of simulated pulmonary nodules of using
  three standard ct window settings.
\newblock \emph{Clinical Radiology}, 47\penalty0 (4):\penalty0 241--244, 1993.
\newblock ISSN 0009-9260.
\newblock \doi{https://doi.org/10.1016/S0009-9260(05)81130-4}.
\newblock URL
  \url{https://www.sciencedirect.com/science/article/pii/S0009926005811304}.

\bibitem[Hase and Bansal(2020)]{hase-bansal-2020-evaluating}
P.~Hase and M.~Bansal.
\newblock Evaluating explainable {AI}: Which algorithmic explanations help
  users predict model behavior?
\newblock In \emph{Proceedings of the 58th Annual Meeting of the Association
  for Computational Linguistics}, pages 5540--5552, Online, July 2020.
  Association for Computational Linguistics.
\newblock \doi{10.18653/v1/2020.acl-main.491}.
\newblock URL \url{https://aclanthology.org/2020.acl-main.491}.

\bibitem[He et~al.(2019)He, Baxter, Xu, Xu, Zhou, and Zhang]{He2019}
J.~He, S.~L. Baxter, J.~Xu, J.~Xu, X.~Zhou, and K.~Zhang.
\newblock The practical implementation of artificial intelligence technologies
  in medicine.
\newblock \emph{Nature Medicine}, 25\penalty0 (1):\penalty0 30--36, Jan. 2019.
\newblock \doi{10.1038/s41591-018-0307-0}.
\newblock URL \url{https://doi.org/10.1038/s41591-018-0307-0}.

\bibitem[ho~Cho et~al.(2018)ho~Cho, hak Lee, Kim, and Park]{Cho2018}
H.~ho~Cho, S.~hak Lee, J.~Kim, and H.~Park.
\newblock Classification of the glioma grading using radiomics analysis.
\newblock \emph{{PeerJ}}, 6:\penalty0 e5982, Nov. 2018.
\newblock \doi{10.7717/peerj.5982}.
\newblock URL \url{https://doi.org/10.7717/peerj.5982}.

\bibitem[Hooker et~al.(2019)Hooker, Erhan, Kindermans, and
  Kim]{DBLP:conf/nips/HookerEKK19}
S.~Hooker, D.~Erhan, P.-J. Kindermans, and B.~Kim.
\newblock A benchmark for interpretability methods in deep neural networks.
\newblock In \emph{NeurIPS}, pages 9734--9745, 2019.
\newblock URL
  \url{http://papers.nips.cc/paper/9167-a-benchmark-for-interpretability-methods-in-deep-neural-networks}.

\bibitem[Huang et~al.(2017)Huang, Liu, Van Der~Maaten, and Weinberger]{8099726}
G.~Huang, Z.~Liu, L.~Van Der~Maaten, and K.~Q. Weinberger.
\newblock Densely connected convolutional networks.
\newblock In \emph{2017 IEEE Conference on Computer Vision and Pattern
  Recognition (CVPR)}, pages 2261--2269, 2017.
\newblock \doi{10.1109/CVPR.2017.243}.

\bibitem[Jacovi and Goldberg(2020)]{jacovi-goldberg-2020-towards}
A.~Jacovi and Y.~Goldberg.
\newblock Towards faithfully interpretable {NLP} systems: How should we define
  and evaluate faithfulness?
\newblock In \emph{Proceedings of the 58th Annual Meeting of the Association
  for Computational Linguistics}, pages 4198--4205, Online, July 2020.
  Association for Computational Linguistics.
\newblock \doi{10.18653/v1/2020.acl-main.386}.
\newblock URL \url{https://aclanthology.org/2020.acl-main.386}.

\bibitem[Jin and Hamarneh()]{jin-doctor-user-study}
W.~Jin and G.~Hamarneh.
\newblock What explanations do doctors require from artificial intelligence?
\newblock In \emph{Manuscript in preparation}.

\bibitem[Jin et~al.(2020)Jin, Fatehi, Abhishek, Mallya, Toyota, and
  Hamarneh]{Jin_2020}
W.~Jin, M.~Fatehi, K.~Abhishek, M.~Mallya, B.~Toyota, and G.~Hamarneh.
\newblock {Artificial intelligence in glioma imaging: challenges and advances}.
\newblock \emph{Journal of Neural Engineering}, 17\penalty0 (2):\penalty0
  21002, apr 2020.
\newblock \doi{10.1088/1741-2552/ab8131}.

\bibitem[Jin et~al.(2021{\natexlab{a}})Jin, Fan, Gromala, Pasquier, and
  Hamarneh]{jin2021euca}
W.~Jin, J.~Fan, D.~Gromala, P.~Pasquier, and G.~Hamarneh.
\newblock {EUCA}: the end-user-centered explainable {AI} framework,
  2021{\natexlab{a}}.

\bibitem[Jin et~al.(2021{\natexlab{b}})Jin, Li, and
  Hamarneh]{DBLP:journals/corr/abs-2107-05047}
W.~Jin, X.~Li, and G.~Hamarneh.
\newblock One map does not fit all: Evaluating saliency map explanation on
  multi-modal medical images.
\newblock \emph{CoRR}, abs/2107.05047, 2021{\natexlab{b}}.
\newblock URL \url{https://arxiv.org/abs/2107.05047}.

\bibitem[Jin et~al.(2022)Jin, Li, and Hamarneh]{aaai2022}
W.~Jin, X.~Li, and G.~Hamarneh.
\newblock Evaluating explainable {AI} on a multi-modal medical imaging task:
  Can existing algorithms fulfill clinical requirements?
\newblock \emph{Proceedings of the AAAI Conference on Artificial Intelligence},
  36\penalty0 (11):\penalty0 11945--11953, Jun. 2022.
\newblock \doi{10.1609/aaai.v36i11.21452}.
\newblock URL \url{https://ojs.aaai.org/index.php/AAAI/article/view/21452}.

\bibitem[Kawahara et~al.(2019)Kawahara, Daneshvar, Argenziano, and
  Hamarneh]{8333693}
J.~Kawahara, S.~Daneshvar, G.~Argenziano, and G.~Hamarneh.
\newblock Seven-point checklist and skin lesion classification using multitask
  multimodal neural nets.
\newblock \emph{IEEE Journal of Biomedical and Health Informatics}, 23\penalty0
  (2):\penalty0 538--546, 2019.
\newblock \doi{10.1109/JBHI.2018.2824327}.

\bibitem[Kelly et~al.(2019)Kelly, Karthikesalingam, Suleyman, Corrado, and
  King]{Kelly2019}
C.~J. Kelly, A.~Karthikesalingam, M.~Suleyman, G.~Corrado, and D.~King.
\newblock {Key challenges for delivering clinical impact with artificial
  intelligence}.
\newblock \emph{BMC Medicine}, 17\penalty0 (1):\penalty0 195, 2019.
\newblock ISSN 1741-7015.
\newblock \doi{10.1186/s12916-019-1426-2}.
\newblock URL \url{https://doi.org/10.1186/s12916-019-1426-2}.

\bibitem[Kim et~al.(2018)Kim, Wattenberg, Gilmer, Cai, Wexler, Viegas, and
  sayres]{pmlr-v80-kim18d}
B.~Kim, M.~Wattenberg, J.~Gilmer, C.~Cai, J.~Wexler, F.~Viegas, and R.~sayres.
\newblock Interpretability beyond feature attribution: Quantitative testing
  with concept activation vectors ({TCAV}).
\newblock In J.~Dy and A.~Krause, editors, \emph{Proceedings of the 35th
  International Conference on Machine Learning}, volume~80 of \emph{Proceedings
  of Machine Learning Research}, pages 2668--2677. PMLR, 10--15 Jul 2018.
\newblock URL \url{http://proceedings.mlr.press/v80/kim18d.html}.

\bibitem[Kim et~al.(2021)Kim, Kim, and Park]{Kim2021}
S.~Kim, B.~Kim, and H.~Park.
\newblock Synthesis of brain tumor multicontrast {MR} images for improved data
  augmentation.
\newblock \emph{Medical Physics}, Mar. 2021.
\newblock \doi{10.1002/mp.14701}.
\newblock URL \url{https://doi.org/10.1002/mp.14701}.

\bibitem[Krippendorff(2004)]{krippendorff2004content}
K.~Krippendorff.
\newblock \emph{Content analysis : an introduction to its methodology}.
\newblock Sage, Thousand Oaks, Calif, 2004.
\newblock ISBN 0761915451.

\bibitem[Lagioia(2020)]{gdpr}
G.~Lagioia, Francesca;Sartor.
\newblock {The impact of the General Data Protection Regulation (GDPR) on
  artificial intelligence}.
\newblock 2020.
\newblock \doi{10.2861/293}.
\newblock URL \url{http://www.europarl.europa.eu/thinktank}.

\bibitem[Lansberg et~al.(2000)Lansberg, Albers, Beaulieu, and
  Marks]{Lansberg2000}
M.~G. Lansberg, G.~W. Albers, C.~Beaulieu, and M.~P. Marks.
\newblock Comparison of diffusion-weighted {MRI} and {CT} in acute stroke.
\newblock \emph{Neurology}, 54\penalty0 (8):\penalty0 1557--1561, Apr. 2000.
\newblock \doi{10.1212/wnl.54.8.1557}.
\newblock URL \url{https://doi.org/10.1212/wnl.54.8.1557}.

\bibitem[Li et~al.(2020)Li, Zhou, Dvornek, Gu, Ventola, and
  Duncan]{10.1007/978-3-030-59710-8_77}
X.~Li, Y.~Zhou, N.~C. Dvornek, Y.~Gu, P.~Ventola, and J.~S. Duncan.
\newblock Efficient shapley explanation for features importance estimation
  under uncertainty.
\newblock In A.~L. Martel, P.~Abolmaesumi, D.~Stoyanov, D.~Mateus, M.~A.
  Zuluaga, S.~K. Zhou, D.~Racoceanu, and L.~Joskowicz, editors, \emph{Medical
  Image Computing and Computer Assisted Intervention -- MICCAI 2020}, pages
  792--801, Cham, 2020. Springer International Publishing.
\newblock ISBN 978-3-030-59710-8.

\bibitem[Long et~al.(2020)Long, Moriarty, Cardoen, Gao, Vogl, Jean, Hamarneh,
  and Nabi]{Long2020}
R.~K.~M. Long, K.~P. Moriarty, B.~Cardoen, G.~Gao, A.~W. Vogl, F.~Jean,
  G.~Hamarneh, and I.~R. Nabi.
\newblock Super resolution microscopy and deep learning identify zika virus
  reorganization of the endoplasmic reticulum.
\newblock \emph{Scientific Reports}, 10\penalty0 (1), Dec. 2020.
\newblock \doi{10.1038/s41598-020-77170-3}.
\newblock URL \url{https://doi.org/10.1038/s41598-020-77170-3}.

\bibitem[Lundberg and Lee(2017)]{NIPS2017_8a20a862}
S.~M. Lundberg and S.-I. Lee.
\newblock A unified approach to interpreting model predictions.
\newblock In I.~Guyon, U.~V. Luxburg, S.~Bengio, H.~Wallach, R.~Fergus,
  S.~Vishwanathan, and R.~Garnett, editors, \emph{Advances in Neural
  Information Processing Systems}, volume~30. Curran Associates, Inc., 2017.
\newblock URL
  \url{https://proceedings.neurips.cc/paper/2017/file/8a20a8621978632d76c43dfd28b67767-Paper.pdf}.

\bibitem[Lundberg et~al.(2020)Lundberg, Erion, Chen, DeGrave, Prutkin, Nair,
  Katz, Himmelfarb, Bansal, and Lee]{Lundberg2020}
S.~M. Lundberg, G.~Erion, H.~Chen, A.~DeGrave, J.~M. Prutkin, B.~Nair, R.~Katz,
  J.~Himmelfarb, N.~Bansal, and S.-I. Lee.
\newblock From local explanations to global understanding with explainable {AI}
  for trees.
\newblock \emph{Nature Machine Intelligence}, 2\penalty0 (1):\penalty0 56--67,
  Jan. 2020.
\newblock \doi{10.1038/s42256-019-0138-9}.
\newblock URL \url{https://doi.org/10.1038/s42256-019-0138-9}.

\bibitem[Mann and Whitney(1947)]{10.2307/2236101}
H.~B. Mann and D.~R. Whitney.
\newblock On a test of whether one of two random variables is stochastically
  larger than the other.
\newblock \emph{The Annals of Mathematical Statistics}, 18\penalty0
  (1):\penalty0 50--60, 1947.
\newblock ISSN 00034851.
\newblock URL \url{http://www.jstor.org/stable/2236101}.

\bibitem[Mart{\'{\i}}-Bonmat{\'{\i}} et~al.(2010)Mart{\'{\i}}-Bonmat{\'{\i}},
  Sopena, Bartumeus, and Sopena]{MartBonmat2010}
L.~Mart{\'{\i}}-Bonmat{\'{\i}}, R.~Sopena, P.~Bartumeus, and P.~Sopena.
\newblock Multimodality imaging techniques.
\newblock \emph{Contrast Media {\&} Molecular Imaging}, 5\penalty0
  (4):\penalty0 180--189, July 2010.
\newblock \doi{10.1002/cmmi.393}.
\newblock URL \url{https://doi.org/10.1002/cmmi.393}.

\bibitem[Masic et~al.(2008)Masic, Miokovic, and Muhamedagic]{Masic2008}
I.~Masic, M.~Miokovic, and B.~Muhamedagic.
\newblock Evidence based medicine - new approaches and challenges.
\newblock \emph{Acta Informatica Medica}, 16\penalty0 (4):\penalty0 219, 2008.
\newblock \doi{10.5455/aim.2008.16.219-225}.
\newblock URL \url{https://doi.org/10.5455/aim.2008.16.219-225}.

\bibitem[Mohan et~al.(2020)Mohan, Facciorusso, Khan, Chandan, Kassab,
  Gkolfakis, Tziatzios, Triantafyllou, and Adler]{Mohan2020}
B.~P. Mohan, A.~Facciorusso, S.~R. Khan, S.~Chandan, L.~L. Kassab,
  P.~Gkolfakis, G.~Tziatzios, K.~Triantafyllou, and D.~G. Adler.
\newblock Real-time computer aided colonoscopy versus standard colonoscopy for
  improving adenoma detection rate: A meta-analysis of randomized-controlled
  trials.
\newblock \emph{{EClinicalMedicine}}, 29-30:\penalty0 100622, Dec. 2020.
\newblock \doi{10.1016/j.eclinm.2020.100622}.
\newblock URL \url{https://doi.org/10.1016/j.eclinm.2020.100622}.

\bibitem[Mohseni et~al.(2021)Mohseni, Zarei, and Ragan]{10.1145/3387166}
S.~Mohseni, N.~Zarei, and E.~D. Ragan.
\newblock A multidisciplinary survey and framework for design and evaluation of
  explainable ai systems.
\newblock \emph{ACM Trans. Interact. Intell. Syst.}, 11\penalty0 (3–4), Aug.
  2021.
\newblock ISSN 2160-6455.
\newblock \doi{10.1145/3387166}.
\newblock URL \url{https://doi.org/10.1145/3387166}.

\bibitem[Nan et~al.(2022)Nan, Ser, Walsh, Schönlieb, Roberts, Selby, Howard,
  Owen, Neville, Guiot, Ernst, Pastor, Alberich-Bayarri, Menzel, Walsh, Vos,
  Flerin, Charbonnier, {van Rikxoort}, Chatterjee, Woodruff, Lambin,
  Cerdá-Alberich, Martí-Bonmatí, Herrera, and Yang]{NAN202299}
Y.~Nan, J.~D. Ser, S.~Walsh, C.~Schönlieb, M.~Roberts, I.~Selby, K.~Howard,
  J.~Owen, J.~Neville, J.~Guiot, B.~Ernst, A.~Pastor, A.~Alberich-Bayarri,
  M.~I. Menzel, S.~Walsh, W.~Vos, N.~Flerin, J.-P. Charbonnier, E.~{van
  Rikxoort}, A.~Chatterjee, H.~Woodruff, P.~Lambin, L.~Cerdá-Alberich,
  L.~Martí-Bonmatí, F.~Herrera, and G.~Yang.
\newblock Data harmonisation for information fusion in digital healthcare: A
  state-of-the-art systematic review, meta-analysis and future research
  directions.
\newblock \emph{Information Fusion}, 82:\penalty0 99--122, 2022.
\newblock ISSN 1566-2535.
\newblock \doi{https://doi.org/10.1016/j.inffus.2022.01.001}.
\newblock URL
  \url{https://www.sciencedirect.com/science/article/pii/S156625352200015X}.

\bibitem[Olah et~al.(2017)Olah, Mordvintsev, and Schubert]{Olah2017}
C.~Olah, A.~Mordvintsev, and L.~Schubert.
\newblock Feature visualization.
\newblock \emph{Distill}, 2\penalty0 (11), Nov. 2017.
\newblock \doi{10.23915/distill.00007}.
\newblock URL \url{https://doi.org/10.23915/distill.00007}.

\bibitem[Patel et~al.(2016)Patel, Silverberg, Becker-Weidman, Roth, and
  Deshmukh]{Patel2016}
A.~Patel, C.~Silverberg, D.~Becker-Weidman, C.~Roth, and S.~Deshmukh.
\newblock Understanding body {MRI} sequences and their ability to characterize
  tissues.
\newblock \emph{Universal Journal of Medical Science}, 4\penalty0 (1):\penalty0
  1--9, Jan. 2016.
\newblock \doi{10.13189/ujmsj.2016.040101}.
\newblock URL \url{https://doi.org/10.13189/ujmsj.2016.040101}.

\bibitem[Patro et~al.(2019)Patro, Lunayach, Patel, and
  Namboodiri]{Patro_2019_ICCV}
B.~N. Patro, M.~Lunayach, S.~Patel, and V.~P. Namboodiri.
\newblock U-cam: Visual explanation using uncertainty based class activation
  maps.
\newblock In \emph{Proceedings of the IEEE/CVF International Conference on
  Computer Vision (ICCV)}, October 2019.

\bibitem[Pereira et~al.(2018)Pereira, Meier, McKinley, Wiest, Alves, Silva, and
  Reyes]{PEREIRA2018228}
S.~Pereira, R.~Meier, R.~McKinley, R.~Wiest, V.~Alves, C.~A. Silva, and
  M.~Reyes.
\newblock Enhancing interpretability of automatically extracted machine
  learning features: application to a rbm-random forest system on brain lesion
  segmentation.
\newblock \emph{Medical Image Analysis}, 44:\penalty0 228--244, 2018.
\newblock ISSN 1361-8415.
\newblock \doi{https://doi.org/10.1016/j.media.2017.12.009}.
\newblock URL
  \url{https://www.sciencedirect.com/science/article/pii/S1361841517301901}.

\bibitem[Rajpurkar et~al.(2022)Rajpurkar, Chen, Banerjee, and
  Topol]{Rajpurkar2022}
P.~Rajpurkar, E.~Chen, O.~Banerjee, and E.~J. Topol.
\newblock {AI} in health and medicine.
\newblock \emph{Nature Medicine}, 28\penalty0 (1):\penalty0 31--38, Jan. 2022.
\newblock \doi{10.1038/s41591-021-01614-0}.
\newblock URL \url{https://doi.org/10.1038/s41591-021-01614-0}.

\bibitem[Ray(2017)]{Ray2017}
K.~Ray.
\newblock Modelling human stomach development with gastric organoids.
\newblock \emph{Nature Reviews Gastroenterology {\&} Hepatology}, 14\penalty0
  (2):\penalty0 68--68, Jan. 2017.
\newblock \doi{10.1038/nrgastro.2017.4}.
\newblock URL \url{https://doi.org/10.1038/nrgastro.2017.4}.

\bibitem[Ren et~al.(2021)Ren, Zhou, Chen, and
  Zhang]{DBLP:journals/corr/abs-2105-10719}
J.~Ren, Z.~Zhou, Q.~Chen, and Q.~Zhang.
\newblock Learning baseline values for shapley values.
\newblock \emph{CoRR}, abs/2105.10719, 2021.
\newblock URL \url{https://arxiv.org/abs/2105.10719}.

\bibitem[Ribeiro et~al.(2016)Ribeiro, Singh, and Guestrin]{Ribeiro2016b}
M.~T. Ribeiro, S.~Singh, and C.~Guestrin.
\newblock {"Why Should I Trust You?": Explaining the Predictions of Any
  Classifier}.
\newblock In \emph{Proceedings of the 22nd ACM SIGKDD International Conference
  on Knowledge Discovery and Data Mining - KDD '16}, pages 1135--1144, New
  York, New York, USA, 2016. ACM Press.
\newblock ISBN 9781450342322.
\newblock \doi{10.1145/2939672.2939778}.
\newblock URL \url{http://dl.acm.org/citation.cfm?doid=2939672.2939778}.

\bibitem[Rosas and Smet(2009)]{Rosas2009}
H.~G. Rosas and A.~A.~D. Smet.
\newblock Magnetic resonance imaging of the meniscus.
\newblock \emph{Topics in Magnetic Resonance Imaging}, 20\penalty0
  (3):\penalty0 151--173, June 2009.
\newblock \doi{10.1097/rmr.0b013e3181d657d1}.
\newblock URL \url{https://doi.org/10.1097/rmr.0b013e3181d657d1}.

\bibitem[Rudin(2019)]{Rudin2019}
C.~Rudin.
\newblock Stop explaining black box machine learning models for high stakes
  decisions and use interpretable models instead.
\newblock \emph{Nature Machine Intelligence}, 1\penalty0 (5):\penalty0
  206--215, May 2019.
\newblock \doi{10.1038/s42256-019-0048-x}.
\newblock URL \url{https://doi.org/10.1038/s42256-019-0048-x}.

\bibitem[Sackett et~al.(1996)Sackett, Rosenberg, Gray, Haynes, and
  Richardson]{Sackett71}
D.~L. Sackett, W.~M.~C. Rosenberg, J.~A.~M. Gray, R.~B. Haynes, and W.~S.
  Richardson.
\newblock Evidence based medicine: what it is and what it
  isn{\textquoteright}t.
\newblock \emph{BMJ}, 312\penalty0 (7023):\penalty0 71--72, 1996.
\newblock ISSN 0959-8138.
\newblock \doi{10.1136/bmj.312.7023.71}.
\newblock URL \url{https://www.bmj.com/content/312/7023/71}.

\bibitem[{Samek} et~al.(2017){Samek}, {Binder}, {Montavon}, {Lapuschkin}, and
  {Müller}]{7552539}
W.~{Samek}, A.~{Binder}, G.~{Montavon}, S.~{Lapuschkin}, and K.~{Müller}.
\newblock Evaluating the visualization of what a deep neural network has
  learned.
\newblock \emph{IEEE Transactions on Neural Networks and Learning Systems},
  28\penalty0 (11):\penalty0 2660--2673, 2017.
\newblock \doi{10.1109/TNNLS.2016.2599820}.

\bibitem[Saporta et~al.(2021)Saporta, Gui, Agrawal, Pareek, Truong, Nguyen,
  Ngo, Seekins, Blankenberg, Ng, Lungren, and
  Rajpurkar]{Saporta2021.02.28.21252634}
A.~Saporta, X.~Gui, A.~Agrawal, A.~Pareek, S.~Q. Truong, C.~D. Nguyen, V.-D.
  Ngo, J.~Seekins, F.~G. Blankenberg, A.~Y. Ng, M.~P. Lungren, and
  P.~Rajpurkar.
\newblock Deep learning saliency maps do not accurately highlight
  diagnostically relevant regions for medical image interpretation.
\newblock \emph{medRxiv}, 2021.
\newblock \doi{10.1101/2021.02.28.21252634}.
\newblock URL
  \url{https://www.medrxiv.org/content/early/2021/03/02/2021.02.28.21252634}.

\bibitem[{Selvaraju} et~al.(2017){Selvaraju}, {Cogswell}, {Das}, {Vedantam},
  {Parikh}, and {Batra}]{8237336}
R.~R. {Selvaraju}, M.~{Cogswell}, A.~{Das}, R.~{Vedantam}, D.~{Parikh}, and
  D.~{Batra}.
\newblock Grad-cam: Visual explanations from deep networks via gradient-based
  localization.
\newblock In \emph{2017 IEEE International Conference on Computer Vision
  (ICCV)}, pages 618--626, 2017.
\newblock \doi{10.1109/ICCV.2017.74}.

\bibitem[Shapley(1951)]{RM-670-PR}
L.~S. Shapley.
\newblock \emph{Notes on the n-Person Game -- II: The Value of an n-Person
  Game}.
\newblock RAND Corporation, Santa Monica, CA, 1951.

\bibitem[Shrikumar et~al.(2017{\natexlab{a}})Shrikumar, Greenside, and
  Kundaje]{10.5555/3305890.3306006}
A.~Shrikumar, P.~Greenside, and A.~Kundaje.
\newblock Learning important features through propagating activation
  differences.
\newblock In \emph{Proceedings of the 34th International Conference on Machine
  Learning - Volume 70}, ICML'17, page 3145–3153. JMLR.org,
  2017{\natexlab{a}}.

\bibitem[Shrikumar et~al.(2017{\natexlab{b}})Shrikumar, Greenside, Shcherbina,
  and Kundaje]{shrikumar2017just}
A.~Shrikumar, P.~Greenside, A.~Shcherbina, and A.~Kundaje.
\newblock Not just a black box: Learning important features through propagating
  activation differences, 2017{\natexlab{b}}.

\bibitem[Simonyan and Zisserman(2015)]{DBLP:journals/corr/SimonyanZ14a}
K.~Simonyan and A.~Zisserman.
\newblock Very deep convolutional networks for large-scale image recognition.
\newblock In Y.~Bengio and Y.~LeCun, editors, \emph{3rd International
  Conference on Learning Representations, {ICLR} 2015, San Diego, CA, USA, May
  7-9, 2015, Conference Track Proceedings}, 2015.
\newblock URL \url{http://arxiv.org/abs/1409.1556}.

\bibitem[Simonyan et~al.(2014)Simonyan, Vedaldi, and
  Zisserman]{simonyan2014deep}
K.~Simonyan, A.~Vedaldi, and A.~Zisserman.
\newblock Deep inside convolutional networks: Visualising image classification
  models and saliency maps, 2014.

\bibitem[Singh et~al.(2020{\natexlab{a}})Singh, Sengupta, J., Mohammed, Faruq,
  Jayakumar, Zelek, and Lakshminarayanan]{10.1007/978-3-030-63419-3_3}
A.~Singh, S.~Sengupta, J.~B. J., A.~R. Mohammed, I.~Faruq, V.~Jayakumar,
  J.~Zelek, and V.~Lakshminarayanan.
\newblock What is the optimal attribution method for explainable ophthalmic
  disease classification?
\newblock In H.~Fu, M.~K. Garvin, T.~MacGillivray, Y.~Xu, and Y.~Zheng,
  editors, \emph{Ophthalmic Medical Image Analysis}, pages 21--31, Cham,
  2020{\natexlab{a}}. Springer International Publishing.
\newblock ISBN 978-3-030-63419-3.

\bibitem[Singh et~al.(2020{\natexlab{b}})Singh, Sengupta, and
  Lakshminarayanan]{Singh2020}
A.~Singh, S.~Sengupta, and V.~Lakshminarayanan.
\newblock Explainable deep learning models in medical image analysis.
\newblock \emph{Journal of Imaging}, 6\penalty0 (6):\penalty0 52, June
  2020{\natexlab{b}}.
\newblock \doi{10.3390/jimaging6060052}.
\newblock URL \url{https://doi.org/10.3390/jimaging6060052}.

\bibitem[Slack et~al.(2021)Slack, Hilgard, Singh, and
  Lakkaraju]{NEURIPS2021_4e246a38}
D.~Slack, A.~Hilgard, S.~Singh, and H.~Lakkaraju.
\newblock Reliable post hoc explanations: Modeling uncertainty in
  explainability.
\newblock In M.~Ranzato, A.~Beygelzimer, Y.~Dauphin, P.~Liang, and J.~W.
  Vaughan, editors, \emph{Advances in Neural Information Processing Systems},
  volume~34, pages 9391--9404. Curran Associates, Inc., 2021.
\newblock URL
  \url{https://proceedings.neurips.cc/paper/2021/file/4e246a381baf2ce038b3b0f82c7d6fb4-Paper.pdf}.

\bibitem[Smilkov et~al.(2017)Smilkov, Thorat, Kim, Viégas, and
  Wattenberg]{smilkov2017smoothgrad}
D.~Smilkov, N.~Thorat, B.~Kim, F.~Viégas, and M.~Wattenberg.
\newblock Smoothgrad: removing noise by adding noise, 2017.

\bibitem[Sokol and Flach(2020)]{Sokol2020}
K.~Sokol and P.~Flach.
\newblock {Explainability fact sheets: A framework for systematic assessment of
  explainable approaches}.
\newblock \emph{FAT* 2020 - Proceedings of the 2020 Conference on Fairness,
  Accountability, and Transparency}, pages 56--67, 2020.
\newblock \doi{10.1145/3351095.3372870}.

\bibitem[Song et~al.(2013)Song, Treanor, Bulpitt, and Magee]{Song2013}
Y.~Song, D.~Treanor, A.~Bulpitt, and D.~Magee.
\newblock 3d reconstruction of multiple stained histology images.
\newblock \emph{Journal of Pathology Informatics}, 4\penalty0 (2):\penalty0 7,
  2013.
\newblock \doi{10.4103/2153-3539.109864}.
\newblock URL \url{https://doi.org/10.4103/2153-3539.109864}.

\bibitem[Springenberg et~al.(2015)Springenberg, Dosovitskiy, Brox, and
  Riedmiller]{springenberg2015striving}
J.~T. Springenberg, A.~Dosovitskiy, T.~Brox, and M.~Riedmiller.
\newblock Striving for simplicity: The all convolutional net, 2015.

\bibitem[Sundararajan et~al.(2017)Sundararajan, Taly, and
  Yan]{10.5555/3305890.3306024}
M.~Sundararajan, A.~Taly, and Q.~Yan.
\newblock Axiomatic attribution for deep networks.
\newblock In \emph{Proceedings of the 34th International Conference on Machine
  Learning - Volume 70}, ICML'17, page 3319–3328. JMLR.org, 2017.

\bibitem[Taghanaki et~al.(2019)Taghanaki, Havaei, Berthier, Dutil, Di~Jorio,
  Hamarneh, and Bengio]{10.1007/978-3-030-32226-7_82}
S.~A. Taghanaki, M.~Havaei, T.~Berthier, F.~Dutil, L.~Di~Jorio, G.~Hamarneh,
  and Y.~Bengio.
\newblock Infomask: Masked variational latent representation to localize chest
  disease.
\newblock In D.~Shen, T.~Liu, T.~M. Peters, L.~H. Staib, C.~Essert, S.~Zhou,
  P.-T. Yap, and A.~Khan, editors, \emph{Medical Image Computing and Computer
  Assisted Intervention -- MICCAI 2019}, pages 739--747, Cham, 2019. Springer
  International Publishing.
\newblock ISBN 978-3-030-32226-7.

\bibitem[Topaloglu et~al.(2021)Topaloglu, Morrell, Rajendran, and
  Topaloglu]{10.3389/frai.2021.746497}
M.~Y. Topaloglu, E.~M. Morrell, S.~Rajendran, and U.~Topaloglu.
\newblock In the pursuit of privacy: The promises and predicaments of federated
  learning in healthcare.
\newblock \emph{Frontiers in Artificial Intelligence}, 4, 2021.
\newblock ISSN 2624-8212.
\newblock \doi{10.3389/frai.2021.746497}.
\newblock URL
  \url{https://www.frontiersin.org/article/10.3389/frai.2021.746497}.

\bibitem[Topol(2019)]{Topol2019}
E.~J. Topol.
\newblock {High-performance medicine: the convergence of human and artificial
  intelligence}.
\newblock \emph{Nature Medicine}, 25\penalty0 (1):\penalty0 44--56, 2019.
\newblock ISSN 1546-170X.
\newblock \doi{10.1038/s41591-018-0300-7}.
\newblock URL \url{https://doi.org/10.1038/s41591-018-0300-7}.

\bibitem[Vilone and Longo(2021)]{VILONE202189}
G.~Vilone and L.~Longo.
\newblock Notions of explainability and evaluation approaches for explainable
  artificial intelligence.
\newblock \emph{Information Fusion}, 76:\penalty0 89--106, 2021.
\newblock ISSN 1566-2535.
\newblock \doi{https://doi.org/10.1016/j.inffus.2021.05.009}.
\newblock URL
  \url{https://www.sciencedirect.com/science/article/pii/S1566253521001093}.

\bibitem[Viviano et~al.(2021)Viviano, Simpson, Dutil, Bengio, and
  Cohen]{viviano2021saliency}
J.~D. Viviano, B.~Simpson, F.~Dutil, Y.~Bengio, and J.~P. Cohen.
\newblock Saliency is a possible red herring when diagnosing poor
  generalization.
\newblock In \emph{International Conference on Learning Representations}, 2021.
\newblock URL \url{https://openreview.net/forum?id=c9-WeM-ceB}.

\bibitem[Wang et~al.(2021)Wang, Wang, Zhang, Wang, Zhu, Gao, Fan, and
  Tian]{2101.01524}
D.~Wang, L.~Wang, Z.~Zhang, D.~Wang, H.~Zhu, Y.~Gao, X.~Fan, and F.~Tian.
\newblock "brilliant ai doctor" in rural china: Tensions and challenges in
  ai-powered cdss deployment.
\newblock 2021.
\newblock \doi{10.1145/3411764.3445432}.

\bibitem[Woo et~al.(2017)Woo, Chang, Lindquist, and Wager]{Woo2017}
C.-W. Woo, L.~J. Chang, M.~A. Lindquist, and T.~D. Wager.
\newblock Building better biomarkers: brain models in translational
  neuroimaging.
\newblock \emph{Nature Neuroscience}, 20\penalty0 (3):\penalty0 365--377, Feb.
  2017.
\newblock \doi{10.1038/nn.4478}.
\newblock URL \url{https://doi.org/10.1038/nn.4478}.

\bibitem[Wu et~al.(2019)Wu, Zhong, Peng, Xu, Huang, Yuan, Ma, and Tan]{Wu2019a}
M.~Wu, X.~Zhong, Q.~Peng, M.~Xu, S.~Huang, J.~Yuan, J.~Ma, and T.~Tan.
\newblock {Prediction of molecular subtypes of breast cancer using BI-RADS
  features based on a “white box” machine learning approach in a
  multi-modal imaging setting}.
\newblock \emph{European Journal of Radiology}, 114:\penalty0 175--184, may
  2019.
\newblock URL
  \url{https://www-sciencedirect-com.proxy.lib.sfu.ca/science/article/pii/S0720048X1930110X{\#}fig0005
  http://www.ncbi.nlm.nih.gov/pubmed/31005170
  https://linkinghub.elsevier.com/retrieve/pii/S0720048X1930110X}.

\bibitem[Xu(2019)]{10.1007/978-3-030-32962-4_18}
Y.~Xu.
\newblock Deep learning in multimodal medical image analysis.
\newblock In H.~Wang, S.~Siuly, R.~Zhou, F.~Martin-Sanchez, Y.~Zhang, and
  Z.~Huang, editors, \emph{Health Information Science}, pages 193--200, Cham,
  2019. Springer International Publishing.
\newblock ISBN 978-3-030-32962-4.

\bibitem[Yang et~al.(2022)Yang, Ye, and Xia]{YANG202229}
G.~Yang, Q.~Ye, and J.~Xia.
\newblock Unbox the black-box for the medical explainable ai via multi-modal
  and multi-centre data fusion: A mini-review, two showcases and beyond.
\newblock \emph{Information Fusion}, 77:\penalty0 29--52, 2022.
\newblock ISSN 1566-2535.
\newblock \doi{https://doi.org/10.1016/j.inffus.2021.07.016}.
\newblock URL
  \url{https://www.sciencedirect.com/science/article/pii/S1566253521001597}.

\bibitem[Ye et~al.(2022)Ye, Gao, Ding, Niu, Wang, Jiang, Wang, Fang,
  Menpes-Smith, Xia, and Yang]{YE2022108291}
Q.~Ye, Y.~Gao, W.~Ding, Z.~Niu, C.~Wang, Y.~Jiang, M.~Wang, E.~F. Fang,
  W.~Menpes-Smith, J.~Xia, and G.~Yang.
\newblock Robust weakly supervised learning for covid-19 recognition using
  multi-center ct images.
\newblock \emph{Applied Soft Computing}, 116:\penalty0 108291, 2022.
\newblock ISSN 1568-4946.
\newblock \doi{https://doi.org/10.1016/j.asoc.2021.108291}.
\newblock URL
  \url{https://www.sciencedirect.com/science/article/pii/S1568494621010966}.

\bibitem[Yeh et~al.(2019)Yeh, Hsieh, Suggala, Inouye, and
  Ravikumar]{NEURIPS2019_a7471fdc}
C.-K. Yeh, C.-Y. Hsieh, A.~Suggala, D.~I. Inouye, and P.~K. Ravikumar.
\newblock On the (in)fidelity and sensitivity of explanations.
\newblock In H.~Wallach, H.~Larochelle, A.~Beygelzimer, F.~d\textquotesingle
  Alch\'{e}-Buc, E.~Fox, and R.~Garnett, editors, \emph{Advances in Neural
  Information Processing Systems}, volume~32. Curran Associates, Inc., 2019.
\newblock URL
  \url{https://proceedings.neurips.cc/paper/2019/file/a7471fdc77b3435276507cc8f2dc2569-Paper.pdf}.

\bibitem[Yin et~al.(2021)Yin, Shi, Hsieh, and
  Chang]{DBLP:journals/corr/abs-2104-08782}
F.~Yin, Z.~Shi, C.~Hsieh, and K.~Chang.
\newblock On the faithfulness measurements for model interpretations.
\newblock \emph{CoRR}, abs/2104.08782, 2021.
\newblock URL \url{https://arxiv.org/abs/2104.08782}.

\bibitem[Zeiler and Fergus(2014)]{10.1007/978-3-319-10590-1_53}
M.~D. Zeiler and R.~Fergus.
\newblock Visualizing and understanding convolutional networks.
\newblock In D.~Fleet, T.~Pajdla, B.~Schiele, and T.~Tuytelaars, editors,
  \emph{Computer Vision -- ECCV 2014}, pages 818--833, Cham, 2014. Springer
  International Publishing.
\newblock ISBN 978-3-319-10590-1.

\bibitem[Zhang and Zhu(2018)]{Zhang2018a}
Q.~Zhang and S.~Zhu.
\newblock {Visual Interpretability for Deep Learning: a Survey}.
\newblock \emph{Frontiers of Information Technology and Electronic
  Engineering}, 19\penalty0 (1):\penalty0 27--39, feb 2018.
\newblock ISSN 20959230.
\newblock \doi{10.1631/FITEE.1700808}.
\newblock URL \url{https://arxiv.org/abs/1802.00614}.

\bibitem[Zhang et~al.(2020)Zhang, {Vera Liao}, and Bellamy]{Zhang2020}
Y.~Zhang, Q.~{Vera Liao}, and R.~K. Bellamy.
\newblock {Efect of confidence and explanation on accuracy and trust
  calibration in AI-assisted decision making}.
\newblock \emph{FAT* 2020 - Proceedings of the 2020 Conference on Fairness,
  Accountability, and Transparency}, pages 295--305, 2020.
\newblock \doi{10.1145/3351095.3372852}.

\bibitem[Zhang et~al.(2019)Zhang, Chen, McGough, Xing, Wang, Bui, Xie, Sapkota,
  Cui, Dhillon, Ahmad, Khalil, Dickinson, Shi, Liu, Su, Cai, and
  Yang]{Zhang2019}
Z.~Zhang, P.~Chen, M.~McGough, F.~Xing, C.~Wang, M.~Bui, Y.~Xie, M.~Sapkota,
  L.~Cui, J.~Dhillon, N.~Ahmad, F.~K. Khalil, S.~I. Dickinson, X.~Shi, F.~Liu,
  H.~Su, J.~Cai, and L.~Yang.
\newblock Pathologist-level interpretable whole-slide cancer diagnosis with
  deep learning.
\newblock \emph{Nature Machine Intelligence}, 1\penalty0 (5):\penalty0
  236--245, May 2019.
\newblock \doi{10.1038/s42256-019-0052-1}.
\newblock URL \url{https://doi.org/10.1038/s42256-019-0052-1}.

\bibitem[Zherebtsov et~al.(2019)Zherebtsov, Dremin, Popov, Doronin, Kurakina,
  Kirillin, Meglinski, and Bykov]{Zherebtsov2019}
E.~Zherebtsov, V.~Dremin, A.~Popov, A.~Doronin, D.~Kurakina, M.~Kirillin,
  I.~Meglinski, and A.~Bykov.
\newblock Hyperspectral imaging of human skin aided by artificial neural
  networks.
\newblock \emph{Biomedical Optics Express}, 10\penalty0 (7):\penalty0 3545,
  June 2019.
\newblock \doi{10.1364/boe.10.003545}.
\newblock URL \url{https://doi.org/10.1364/boe.10.003545}.

\bibitem[Zhou et~al.(2021)Zhou, Booth, Ribeiro, and
  Shah]{DBLP:journals/corr/abs-2104-14403}
Y.~Zhou, S.~Booth, M.~T. Ribeiro, and J.~Shah.
\newblock Do feature attribution methods correctly attribute features?
\newblock \emph{CoRR}, abs/2104.14403, 2021.
\newblock URL \url{https://arxiv.org/abs/2104.14403}.

\bibitem[Zintgraf et~al.(2017)Zintgraf, Cohen, Adel, and
  Welling]{DBLP:conf/iclr/ZintgrafCAW17}
L.~M. Zintgraf, T.~S. Cohen, T.~Adel, and M.~Welling.
\newblock Visualizing deep neural network decisions: Prediction difference
  analysis.
\newblock In \emph{5th International Conference on Learning Representations,
  {ICLR} 2017, Toulon, France, April 24-26, 2017, Conference Track
  Proceedings}. OpenReview.net, 2017.
\newblock URL \url{https://openreview.net/forum?id=BJ5UeU9xx}.

\end{thebibliography}
\clearpage

\end{document}